\documentclass{article} 
\usepackage{iclr2025_conference,times}

\usepackage{hyperref}
\usepackage{url}
\usepackage{wrapfig}
\usepackage{caption}
\usepackage{subcaption}
\usepackage{algorithm}
\usepackage{algpseudocode}
\usepackage{svg}
\usepackage{hyperref}
\usepackage{url}
\usepackage{graphicx}
\usepackage{booktabs}
\usepackage{amssymb}
\usepackage{stmaryrd}
\usepackage{subcaption}
\usepackage{wrapfig}
\usepackage{makecell}

\usepackage{amsthm}
\newtheorem{lemma}{Lemma}
\definecolor{attrcolor}{rgb}{0, 0.0, 1.0}
\newcommand{\clr}[1]{{\color{attrcolor}#1}}

\usepackage{amsmath,amsfonts,bm}

\def\eqref#1{equation~\ref{#1}}

\def\1{\bm{1}}

\DeclareMathAlphabet{\mathsfit}{\encodingdefault}{\sfdefault}{m}{sl}
\SetMathAlphabet{\mathsfit}{bold}{\encodingdefault}{\sfdefault}{bx}{n}

\title{Grounding Continuous Representations \\ in Geometry: Equivariant Neural Fields}

\iclrfinalcopy
\author{%
David R. Wessels$^{*, 1}$,
David M. Knigge$^{*, 2}$,
Riccardo Valperga$^{2}$,
Samuele Papa$^{2}$,\\
\textbf{Sharvaree Vadgama$^{1}$,
Efstratios Gavves $^{2}$,
Erik J. Bekkers $^{1}$}\\
$^{1}$AMLab, $^{2}$VISLab, University of Amsterdam \\
\texttt{d.r.wessels@uva.nl},
\texttt{d.m.knigge@uva.nl}  \\
}

\begin{document}

\maketitle

\begin{abstract}
\textit{Conditional Neural Fields} (CNFs) are increasingly being leveraged as continuous signal representations, by associating each data-sample with a latent variable that conditions a shared backbone Neural Field (NeF) to reconstruct the sample. However, existing CNF architectures face limitations when using this latent \textit{downstream} in tasks requiring fine-grained geometric reasoning, such as classification and segmentation. We posit that this results from lack of explicit modelling of geometric information (e.g. locality in the signal or the orientation of a feature) in the latent space of CNFs. As such, we propose Equivariant Neural Fields (ENFs), a novel CNF architecture which uses a geometry-informed cross-attention to condition the NeF on a geometric variable—a latent point cloud of features—that enables an \textit{equivariant} decoding from latent to field. We show that this approach induces a \textit{steerability} property by which both field and latent are grounded in geometry and amenable to transformation laws: if the field transforms, the latent representation transforms accordingly—and vice versa. Crucially, this equivariance relation ensures that the latent is capable of (1) \textit{representing geometric patterns faitfhully}, allowing for geometric reasoning in latent space, (2) \textit{weight-sharing over similar local patterns}, allowing for efficient learning of datasets of fields. We validate these main properties in a range of tasks including classification, segmentation, forecasting, reconstruction and generative modelling, showing clear improvement over baselines with a geometry-free latent space. \textit{Code attached to submission } \href{https://github.com/Dafidofff/enf-jax}{here}. \textit{Code for a clean and minimal repo } \href{https://github.com/david-knigge/enf-min-jax}{here}.
\end{abstract}

\section{Introduction}
{\let\thefootnote\relax\footnotetext{\hspace{-5mm}* Equal contribution.}}
Neural Fields (NeFs) \citep{xie2022neural} have recently gained prominence in the machine learning community as a novel representation method that models data as continuous functions. These fields, expressed as \( f_\theta:\mathbb{R}^d \rightarrow \mathbb{R}^c \), map spatial coordinates—such as pixel locations \( x \in \mathbb{R}^2 \)—to a corresponding signal, like RGB values \( f_\theta(x) \in \mathbb{R}^3 \), with \( \theta \) representing the model's parameters. The parameters are optimized to approximate a target signal \( f \), ensuring \( \forall x: f(x) \approx f_\theta(x) \). This capability makes NeFs effective for representing continuous spatial, spatio-temporal, and geometric data, particularly in cases where grid-based methods fall short \citep{dupont2022data}. Their promise lies in serving as resolution-free representation that may be used irrespective of data resolution or discretization \citep{xie2022neural}. Moreover, NeF-representations unify downstream models over different data modalities, allowing for transfer of modelling principles between data modalities that otherwise require data-specific engineering efforts \citep{dupont2022data,papa2023train}.

Building on this concept, \textit{conditional neural fields} (CNFs) introduce a conditioning variable \( z \in \mathcal{Z} \) to the model. Given a dataset of \( N \) fields \( \mathcal{D} = \{f_i:\mathbb{R}^d \rightarrow \mathbb{R}^c\}_{i=1}^N \), each specific field can now be represented by a specific conditioning variable \( z_i \) via \( \forall x: f_i(x) \approx f_\theta(x ; z_i) \), while model weights \( \theta \) are shared over the entire dataset. This approach enables CNFs to efficiently model datasets of fields using a set of latent variables that condition a \textit{shared backbone} NeF. This allows for representing and analysing fields $f_i$ by means of their latent representation $z_i$, enabling novel approaches for solving tasks involving fields through a framework known as \textit{learning with functa} \citep{dupont2022data}. Applications of this include tasks such as classification, segmentation, and the generation of continuous fields \citep{dupont2022data, papa2023train}, as well as continuous PDE forecasting by solving dynamics in the latent space \citep{yin2022continuous, knigge2024space}.

\begin{figure}[t]
    \centering
    \includegraphics[width=\linewidth]{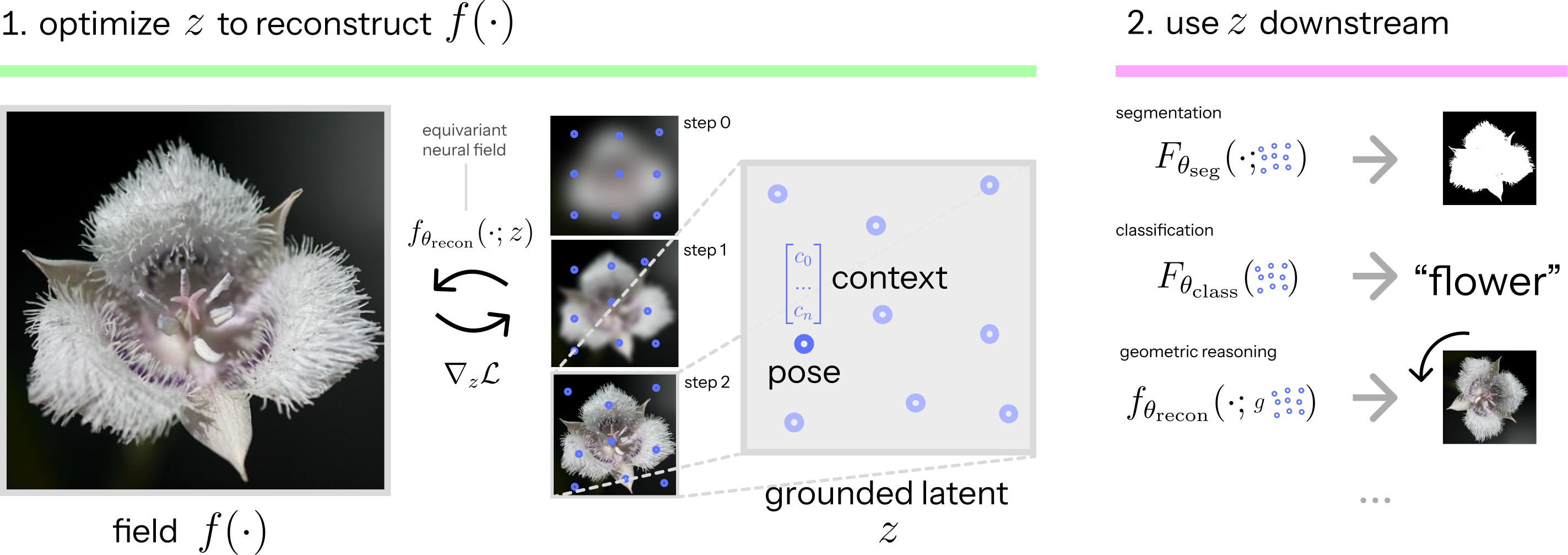}
    \caption{Equivariant Neural Fields (ENFs) ground Neural Fields (NeFs) in geometry using a latent point cloud. A latent set $z$ consisting of tuples $(p_i, \mathbf{c}_i)$ of \textit{pose} information $p_i$ and \textit{context} $\mathbf{c}_i$ is optimized to reconstruct to the field $f(\cdot)$ as a function $f_\theta(\cdot; z)$ using gradient-descent. Due to their explicit positional grounding and locality, the latent retains important geometric features in the input field. The latent $z$ can then be used in downstream tasks, e.g. classification, segmentation, and geometric reasoning, where transformations in the field are mirrored in the latent representation through group actions 
    \( L_g[f] \sim g \cdot z\).
    }
    \label{fig-1}
    \vspace{-5mm}
\end{figure}

\vspace{-2mm}
\paragraph{Geometry in CNFs} A notable limitation of conventional CNFs, however, is a lack of explicit geometric interpretability; each field $f_i$ is encoded by a "global" variable $z_i$, meaning for instance that any notion of locality or other explicit spatial relationships—which have proven a strong inductive bias in computer vision—is lost. Althought this global representation has inherent benefits, e.g. enabling the use of simple MLPs as downstream models and allowing for intuitive interpolation between latent signal representations, empirically it has shown limited performance in settings where samples of the dataset are not consistently globally aligned \citep{bauer2023spatial}. For instance, in classification tasks, spatial organisation of an image's content is crucial for understanding shape and enabling geometric reasoning \citep{van2019capsulenet}; current neural fields lack such geometric inductive biases, limiting their performance in e.g. classification and generative modelling \citep{bauer2023spatial,papa2023train}. To this end, we propose \textit{equivariant neural fields} (ENFs), a new class of NeFs that allows for the identification of continuous fields with concrete geometric representations (Fig. \ref{fig-1}).

\paragraph{Geometry-grounded neural fields} \vspace{-3mm} When the goal is to utilise field representations \( z_i \) in downstream tasks, it is crucial that these representations capture both textural/appearance information and explicit geometric information. Our approach is inspired by the idea of \textit{neural ideograms}—learnable geometric representations \citep{vadgama2022kendall,vadgama2023continuous}, and addresses the pervasive issue of \textit{texture bias}, which causes typical deep learning systems to overfit to textural patterns and ignore important geometric cues \citep{geirhos2018imagenet, hermann2020origins}. To address this challenge, we propose defining representations that (1) explicitly separate aspects of geometry--specifically the \textit{pose} of features--from appearance and (2) are localized in the input signal such that geometric concepts like orientation and distance are maintained from input to latent space. This necessitates that the geometric components of the representations have a meaningful structure, adhering to the same group transformation laws applicable to the fields. Geometrically, this means that distortions in the field translate to corresponding distortions in the latent space, ensuring that geometric (shape) variations are preserved and representable in latent space. 

\begin{figure}
    \centering
    \begin{minipage}{0.49\textwidth}
        \centering
        \includegraphics[width=0.9\textwidth]{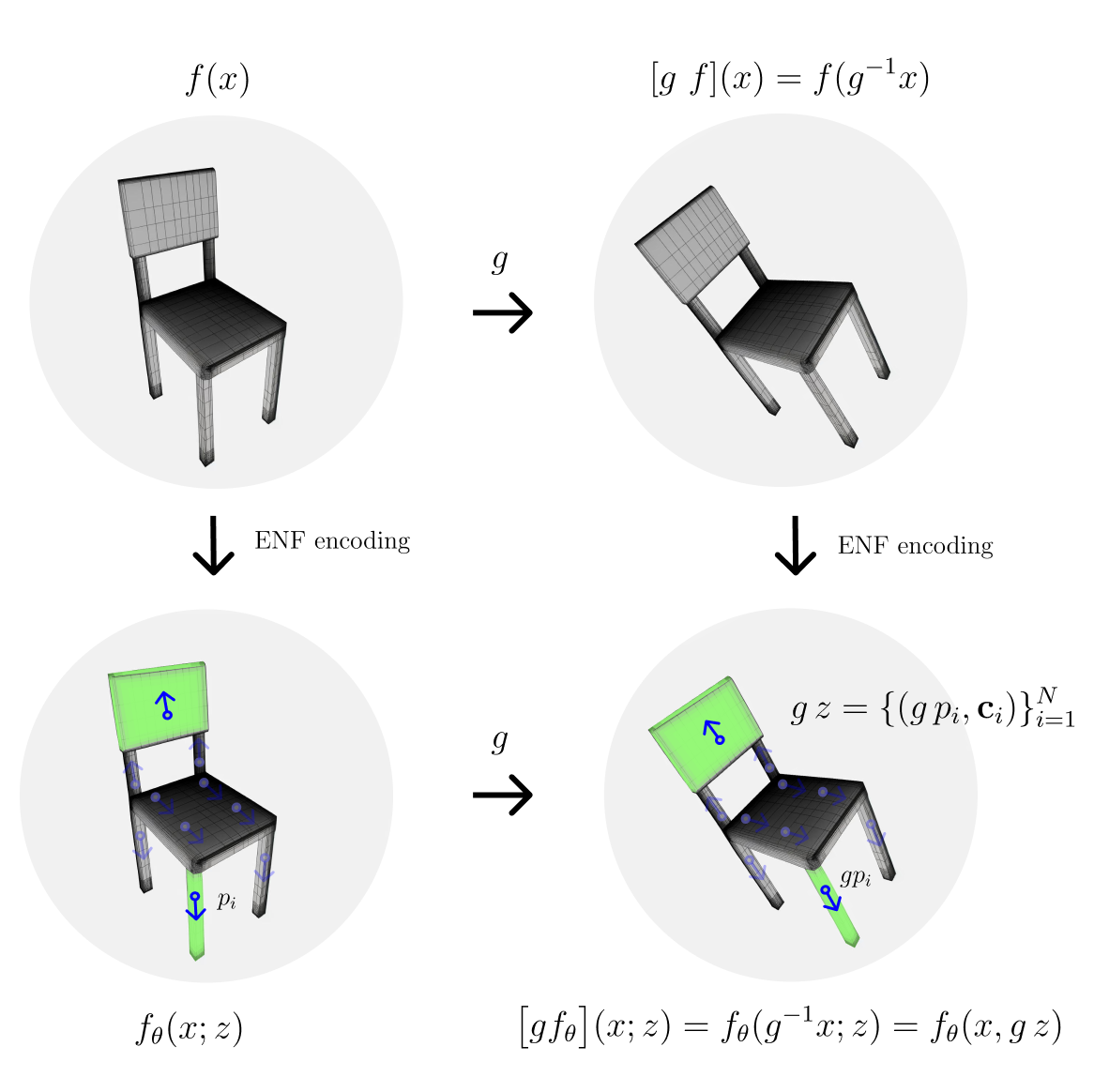}
         \caption{ENFs preserve transformations through their steerability property; if the field transforms with a group action $g$, the latents transform accordingly via the following group action on the pointcloud; $gz=\{gp_i,\mathbf{c}_i\}^N_{i=1}$.}
        \label{fig:steerability}
    \end{minipage}
    \hfill
    \begin{minipage}{0.49\textwidth}
        \centering
        \includegraphics[width=0.9\textwidth]{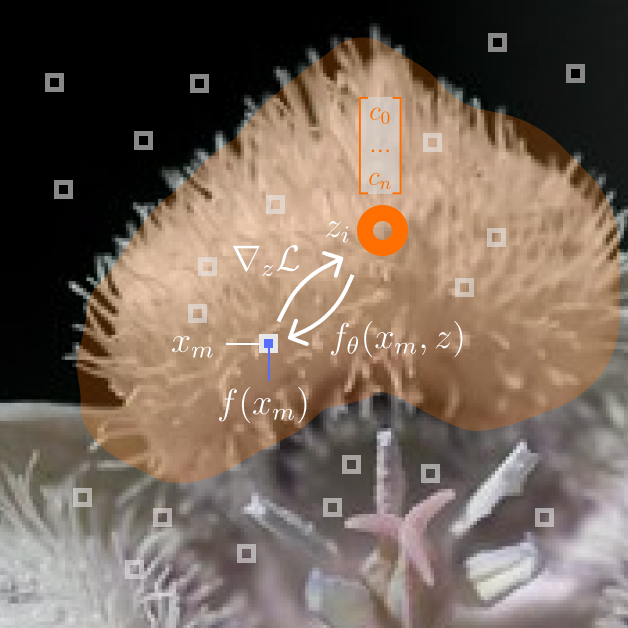}
        \caption{ENFs are a local signal encoding; a latent $z_i$ is optimized to represent a local signal patch. We show that this inductive bias allows for leveraging weight-sharing, and improves downstream performance by retaining important geometric features.}
        \label{fig:locality}
    \end{minipage}

\end{figure}

To establish an explicit grounding in geometry, we propose modelling the conditioning variables as geometric point clouds \( z = \{ (p_i, \mathbf{c}_i) \}_{i=1}^N \), comprising \( N \) pose-appearance tuples, with \( p_i \in G \) being a pose (element) in a group \( G \), and \( \mathbf{c}_i \in \mathbb{R}^c \) an appearance vector. This representation space has a well-defined group action, namely \( g z = \{ (g p_i, \mathbf{c}_i) \}_{i=1}^N \), allowing us to formalise the notion of grounding a neural field through the following 
\begin{equation}
\label{eq:steerability}
\boxed{
 \text{\textit{Steerability property}:}\;\;\;\;\;\;\; \forall g\in G: f_\theta(g^{-1}x; z)=f_\theta(x; gz) \, .}
\end{equation}
This property ensures that if the field transforms, the latent transforms accordingly (Fig. \ref{fig:steerability}). This property has also been shown in \citep{atzmon2022frame, chatzipantazis2022se} for learning implicit shape representations and is well known in equivariance literature more broadly \citep{bekkers2019b,cohen2019general,weiler2019general,deng2021vector}.

\paragraph{Contributions} With this work we present the following contributions: A new class of geometry-grounded equivariant neural fields that posses
\begin{itemize}
    \item  the \textit{steerability} property and thus proveable generalization over group actions
    \item \textit{weight sharing} which enables more efficient learning
    \item \textit{localized representations} in a latent point set which enables unique editing properties
\end{itemize}
We verify these properties through a range of experiments that (1) support the claim that latents are \textit{geometrically meaningful}, (2) show competitive reconstruction and representation capacity on segmentation, classification, forecasting and super-resolution tasks on image and shape data.

\section{Background}
\paragraph{Neural Fields and conditioning variables}
Neural Fields (NeFs) are learned functions $f_\theta$ mapping signal coordinates $x$ to signal values $f(x)$, parameterized by a neural network with parameters $\theta$. Due to their flexibility they have emerged as prominent continuous data representation, applied on datatypes varying from object or scene data \citep{park2019deepsdf, mescheder2019occupancy, sitzmann2020metasdf, mildenhall2021nerf} to audio and images \cite{tancik2020fourier, sitzmann2020implicit}. In order to more efficiently represent whole datasets of signals, Conditional Neural Fields only optimise a latent conditioning variable $z^f$ per signal-instance $f \in D$ instead of optimising a separate set of neural network parameters $\theta_i$. The two most common approaches for obtaining latents $z^f$ are autodecoding \citep{park2019deepsdf} - where latents $z^f$ are initialized and optimized alongside NeF parameters $\theta$ - and Meta-Learning based encoding - where optimization is split into an outer loop that optimizes the backbone $\theta$ and an inner loop that optimizes latents $z^f$. We explain both in detail in Appx. \ref{appx:autodecoding-maml}, and use both in experiments described in Sec. \ref{sec:experiments}.

In seminal work by \cite{dupont2022data}, a datatype agnostic approach for \textit{learning} on these continuous signal representations was proposed - involving first the optimization of a set of conditioning variables $z$ to reconstruct a dataset of signals $\mathcal{D}:= \{f_j\}_{j=1}^n{\sim}\{z^{f_j}\}_{j=1}^n$, and afterward using these variables as a surrogate for the data in downstream tasks such as classification, generation and completion. Although this work highlighted the flexible data-agnostic nature of Conditional Neural Fields (CNFs) by representing signals through a single "global" condition variable $z^{f_j}$, later work by \cite{bauer2023spatial} showed their limitations in performance for more complex tasks (i.e. involving higher-resolution more varied natural data).

\paragraph{Group theoretical preliminaries}
The notion of transformation-preservation of an operator - such as the relation between field $f$ and latent $z$ - is best expressed through group theory. A group is an algebraic construction $(G, \cdot)$ defined by a set $G$ and a binary operator $\cdot: G \times G \rightarrow G$ called the \textit{group product}, satisfying: \textit{closure}: $\forall h,g \in G: h\cdot g \in G$, \textit{identity}: $\exists e \in G: \forall g \in G, g\cdot e=g$, \textit{inverse}: $\forall g \exists g^{-1} \in G: g \cdot g^{-1} = e$ and \textit{associativity}: $\forall g,h,i \in G: (g \cdot h) \cdot i = g \cdot (h \cdot i)$

Given such a group $G$ with identity element $e\in G$, and a set $X$, we can define the \emph{group action} as a map $G \times X \rightarrow X$, which we will denote with direct multiplication i.e. a group element $g\in G$ action on a coordinate $x\in X$ is denoted $gx$. Note that when $X=G$ the group action equals the group product. For the group action on fields $f:X \rightarrow \mathbb{R}$ we use a separate symbol, namely $[\mathcal{L}_g f](x):=f(g^{-1} x)$. In this work we are interested in the Special Euclidean group $SE(n) = T_n \rtimes SO(n)$. $SE(n)$ is the roto-translation group consisting of elements $g = (\mathbf{t}, \mathbf{R})$ with group operation $g\, g' = (\mathbf{t}, \mathbf{R})\,(\mathbf{t'}, \mathbf{R}')=( \mathbf{t} + \mathbf{Rt}', \mathbf{R R}')$; the left-regular action on function spaces is defined by $\mathcal{L}_g f(x) = f(g^{-1}x) = f(\mathbf{R}^{-1}(x-\mathbf{t}))$.

\paragraph{Equivariant graph neural networks for downstream tasks.} \vspace{-2mm}
A key property of our framework is its ability to \textit{associate geometric representations with fields}. This capability unlocks a rich toolset for field analysis through the lens of geometric deep learning (GDL) \citep{bronstein2021geometric}. The GDL field has made significant advancements in the analysis and processing of geometric data, including the study of molecular properties \citep{batzner20223,batatia2022mace,gasteiger2021gemnet,brandstetter2021geometric} and the generation of molecules \citep{hoogeboom2022equivariant,bekkers2023fast} and protein backbones \citep{corso2022diffdock, yim2023fast}. In essence, these approaches \textit{characterise shape}. Our encoding scheme makes these tools now applicable to analysing the geometric components of neural fields, shown in Sec. \ref{sec:experiments}.

Equivariant Graph Neural Networks (EGNNs) are a class of Graph Neural Networks (GNNs) that imposes roto-translational equivariance constraints on their message passing operators to ensure that the learned representations adhere to specific transformation symmetries of the data. 
Among the various forms of equivariant graph NNs \citep{thomas2018tensor,brandstetter2021geometric, satorras2021n,gasteiger2021gemnet,bekkers2023fast,eijkelboom2023n,kofinas2024graph} we will utilise P$\Theta$NITA \citep{bekkers2023fast} as an equivariant operator to analyse and process our latent point-set representations of fields. For the neural field, we leverage the same optimal bi-invariant attributes as introduced in \cite{bekkers2023fast}—which are based on the theory of homogeneous spaces—to parameterise our neural fields, allowing for seamless integration. 
They formalise the notion of \textit{weight sharing} in convolutional networks as the sharing of message functions (kernels) over point-pairs - e.g. relative pixel positions - that should be treated equally. By defining equivalence classes of point-pairs that are identical up to a transformation in the group, we too derive attributes that uniquely identify these classes and enable weight sharing in our proposed ENFs.

\section{Method}\vspace{-2mm}
\label{sec:method}
In this section we introduce the Equivariant Neural Field (ENF) architecture. We start Sec. \ref{sec:operator} by showing how we impose the proposed steerability property (Eq. \ref{eq:steerability}), through the definition of \textit{bi-invariant attributes}. We construct a cross-attention operator conditioned by these attributes. We subsequently constrain the operator to represent local sub-regions of the input domain by applying a Gaussian window. Finally, we propose a k-nearest neighbouring approach to cope with the computational complexity of the cross-attention operation between pixels and latents. In Sec. \ref{subsec:obtaining-z} we finally discuss how we obtain a latent point cloud $z^f$ for a signal $f$.

\paragraph{Bi-invariance constraint} \label{sec:bi-invariance}
Before presenting our equivariant neural field design, we need to understand the constraints imposed by the steerability property (Eq. \ref{eq:steerability}). The key result--shown also by \cite{cohen2019general, bekkers2019b} in the context of equivariant CNNs--is that for steerability, the field $f_\theta$ must be bi-invariant with respect to both coordinates and latents.
\begin{lemma}
A conditional neural field satisfies the steerability property iff it is bi-invariant, i.e., $\forall g \in G:\; f_\theta(g x; g z) = f_\theta(x; z).$
\end{lemma}
\begin{proof}\vspace{-2mm}
If $f_\theta$ satisfies the steerability property, then $f_\theta(g x; g z) = f_\theta(g^{-1} g x; z) = f_\theta(x; z)$, so it is bi-invariant. Conversely, if $f_\theta$ is bi-invariant, then $f_\theta(g^{-1} x; z) = f_\theta(g g^{-1} x; g z) = f_\theta(x; g z)$, satisfying the steerability property \eqref{eq:steerability}.
\end{proof}

\begin{figure}[t]
    \begin{subfigure}[b]{0.329\textwidth}
        \centering
        \includegraphics[width=\textwidth]{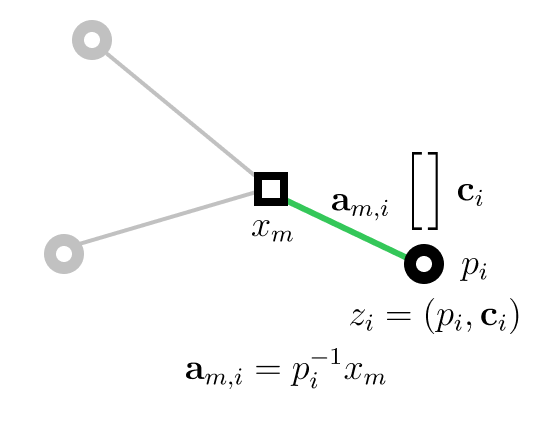}
        \caption{}
        \label{fig:cross-attn-1}
    \end{subfigure}
    \hfill
    \begin{subfigure}{0.329\textwidth}
        \centering
        \includegraphics[width=\textwidth]{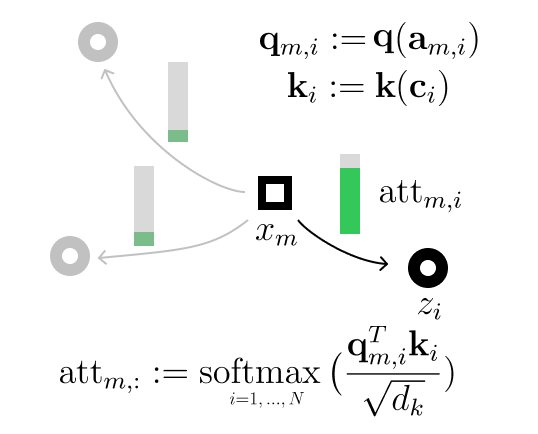}
        \caption{}
        \label{fig:cross-attn-2}
    \end{subfigure}
    \hfill
    \begin{subfigure}{0.329\textwidth}
        \centering
        \includegraphics[width=\textwidth]{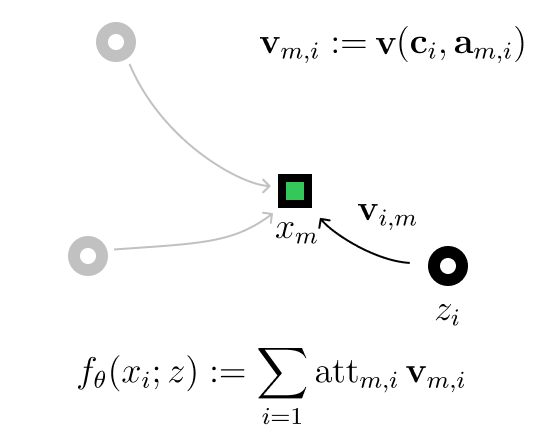}
        \caption{}
        \label{fig:cross-attn-3}
    \end{subfigure}
    \caption{A visual intuition for the proposed cross-attention between coordinate $x_m$ and latent $z=\{(p_i,\mathbf{c}_i)\}_{i=1}^N$. (a) Bi-invariant $\mathbf{a}_{m,i}$ is calculated between coordinate $x_m$ and pose $p_i$ as $p_i^{-1}x_m$. (b) The query and key functions $\mathbf{q}$ transforms $\mathbf{a}_{m,i}$ into a query $\mathbf{q}_{m,i}$, and key function $\mathbf{k}$ maps context vector $\mathbf{c}_i$ to key $\mathbf{k}_i$. Attention coefficients are calculated through a softmax over $\mathbf{q}_{m,i}\mathbf{k}_i$. The softmax is taken over the $N$ latents, yielding $N$ attention coefficients $\operatorname{att}_{m,i}$, one for each latent $z_i$. (c) A value $\mathbf{v}_{m,i}$ for each latent-coordinate pair is calculated as a function $v$ of $\mathbf{c}_i$ and $\mathbf{a}_i$ - and the resulting values are aggregated, weighted by their corresponding attention coefficients $\operatorname{att}_{m,i}$.}
\end{figure}

\subsection{Equivariant Neural Fields conditioned on Geometric Attributes} \label{sec:operator}
The cross-attention operation enables the use of latent-sets as conditioning variables for CNFs by applying cross-attention between embedded coordinates $x_m$ and a latent set of context vectors $z{=}\{ \mathbf{c}_i \}_{i=1}^N$ \citep{zhang20233dshape2vecset}. Such cross-attention fields assign to each $x_m$ a corresponding value $f_\theta(x_m; z)$, by matching a coordinate (query) embedding $\mathbf{q}(x_m)$ against latent (key) vectors $\mathbf{k}(c_i)$ to obtain attention coefficients $\operatorname{att}_{m, i}$, and aggregating associated values $\mathbf{v}(c_i)$ via
$$ \vspace{-1mm}
f_\theta(x_m; z) = \mathbf{W}_o\sum_{i=1}^N \operatorname{att}_{m, i} \mathbf{v}(\mathbf{c}_i) \;\;\;\;\;\;\; \text{with} \;\;\;\;\;\;\; \operatorname{att}_{m, :} = \underset{i{=}1, ..., N}{\operatorname{softmax}} \left( \;\; \frac{\mathbf{q}(x_m)^T \mathbf{k}(\mathbf{c}_i)}{\sqrt{d_k}} \right) \, ,\vspace{-1mm} $$
where $\mathbf{W}_o$ maps cross-attention outputs to NeF output/signal dimension $\mathbb{R}^c$. Our desired latent representation contains a geometric component, namely the poses $p_i$ associated with the context vectors $\mathbf{c}_i$. In order to see how this geometric information could be leveraged, we highlight how each of the three components ($\mathbf{q},\mathbf{k},\mathbf{v}$) could depend on the \clr{geometric attributes}:
$$ \vspace{-1mm}
f_\theta(x_m; z) = \mathbf{W}_o\sum_{i=1}^N \operatorname{att}_{m, i} \mathbf{v}(\clr{x_m}, \clr{p_i}, \mathbf{c}_i) \;\;\;\;\;\;\; \text{with} \;\;\;\;\;\;\; \operatorname{att}_{m, :} = \underset{i{=}1, ..., N}{\operatorname{softmax}} \left( \;\; \frac{\mathbf{q}(\clr{x_m}, \clr{p_i})^T \mathbf{k}(\clr{x_m}, \clr{p_i}, \mathbf{c}_i)}{\sqrt{d_k}} \right) \, . $$
The steerability condition demands that the field has to be bi-invariant with respect to transformations on both $x_m$ and $p_i$, and the easiest way to achieve this is to replace any instance of $x_m, p_i$ by an invariant pair-wise attribute $\clr{\mathbf{a}(x_m, p_i)}$ that is both invariant and maximally informative. With maximally informative we mean that coordinate-pose pairs that are not the same up to a group action receive a different vector descriptor, i.e., $\clr{\mathbf{a}(x_m, p_i)} = \clr{\mathbf{a}(x_m', p_i')}$ \textit{if and only if} there exists a $g \in G$ such that $x_m' = g x_m$ and $p_i' = g p_i$.

\paragraph{Equivariant Neural Fields} Based on recent results in the context of equivariant graph neural networks \cite{bekkers2023fast}, we let $\clr{\mathbf{a}(x_m, p_i):= p_i^{-1} x}$ be the the unique and complete \textit{bijective} identifier for the equivalence class of all coordinate-pose pairs that are the same up to a group action. Bijectivity here implies that the descriptor $\clr{p_i^{-1} x_m}$ contains all information possible to identify the equivalence classes, and thus the use of those attributes leads to maximal expressivity. In this work we use bi-invariants for translation ($\mathbf{a}^{\mathbb{R}^n}$) roto-translation ($\mathbf{a}^{\rm SE(2)}$) - as well as a "bi-invariant" that breaks any equivariance $\mathbf{a}^\emptyset$ (see Appx. \ref{appx:inv_attributes} for details). We define the ENF as follows: 
\vspace{1mm}
\scalebox{0.96}{$$\label{eq:cross-attn}
\boxed{
f_\theta(x; z) = \mathbf{W}_o\sum_{i=1}^N \operatorname{att}_{m, i} \mathbf{v}(\clr{\mathbf{a}(x, p_i)}, \mathbf{c}_i) \;\;\;\;\;\;\; \text{with} \;\;\;\;\;\;\; \operatorname{att}_{m, :}  = \underset{i=1,...,N}{\operatorname{softmax}} \left( \;\; \frac{\mathbf{q}(\clr{\mathbf{a}(x,p_i)})^T \mathbf{k}(\mathbf{c}_i)}{\sqrt{d_k}} \right) \, . 
}
$$
}
\vspace{1mm}
As specific parameterizations for $\mathbf{a},\mathbf{q},\mathbf{k},\mathbf{v},$ we choose:
\begin{align}
\mathbf{a}(x, p_i) &:= \phi(p_i^{-1} x) \\
\mathbf{q}(\mathbf{a}(p_i^{-1}x)) &:= \mathbf{W}_q \clr{\mathbf{a}(x,p_i)} \; \; \; \; \; \; \; \mathbf{k}(\mathbf{c}_i):= \mathbf{W}_k \mathbf{c}_i \, ,\\
\mathbf{v}(\clr{\mathbf{a}(x,p_i)}, \mathbf{c}_i) &:= (\mathbf{W}_v \mathbf{c}_i) \odot (\mathbf{W}_{a_\gamma} \clr{\mathbf{a}(x,p_i)}) + (\mathbf{W}_{a_\beta} \clr{\mathbf{a}(x,p_i)})
\end{align}
with $\odot$ denoting element-wise multiplication and $\phi$ a relative coordinate embedding function which we set to be a Gaussian RFF embedding \citep{tancik2020fourier}. In neural field literature it is known that neural networks suffer from high spectral biases \citep{rahaman2019spectral}. Due to smooth input-output mappings it becomes difficult to learn high-frequency information in low-dimensions such as the coordinate inputs for a NeF. Gaussian RFF embeddings introduce high-frequency signals in the embedding to alleviate the spectral bias.

Since the value transform has as goal to fill in spatially varying signal patches during reconstruction, the value-function is also conditioned on the geometric attributes $p^{-1}x$. To add extra expressivity we chose to apply the conditioning via FiLM modulation \citep{perez2018film} which applies a feature-wise linear modulation with a learnable shift $\beta$ and scale $\gamma$ modulation.

A crucial difference with standard transformer-type methods on point clouds is that cross-attention is between \textit{relative position embeddings}—relative to the latent pose $p_i$—and that the value transform is of depth-wise separable convolutional form \citep{chollet2017xception,bekkers2023fast}, which is a stronger form of conditioning \citep{koishekenov2023exploration} than additive modulation as is typically done in biased self-attention networks such as Point Transformer \citep{zhao2021point}.

\paragraph{Enforcing and learning locality in the latent point cloud}
\label{sec:enforcing_locality}
With the current proposed setup, cross-attention is universally applied across the entire set of latents and coordinates. Given the Softmax distribution, each coordinate indiscriminately receives a nonzero attention value for every latent $(p_i, \mathbf{c}_i)\in z$. Consequently, although latents possess inherent latent space positional attributes, they do not strictly represent localised regions of the signal, breaking the locality we require.

\begin{wrapfigure}[15]{r}{0.5\textwidth}
\vspace{-4mm}
    \begin{minipage}[b]{0.49\linewidth}
        \includegraphics[width=\textwidth]{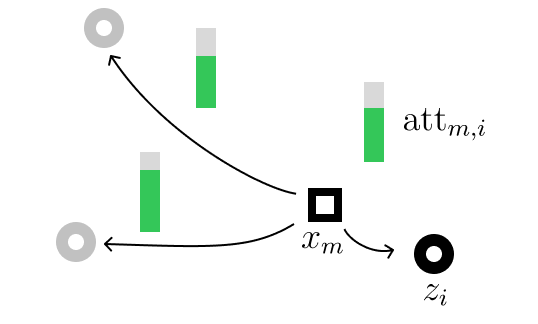}
        \caption*{(a)}
    \end{minipage}
    \begin{minipage}[b]{0.49\linewidth}
        \includegraphics[width=\textwidth]{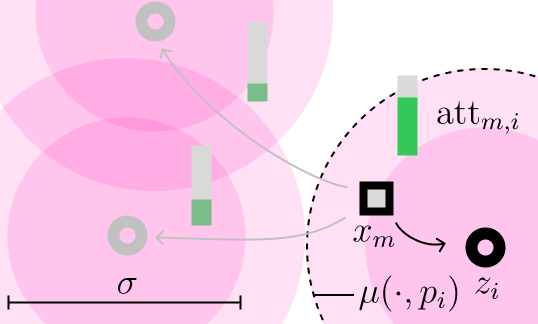}
        \caption*{(b)}
    \end{minipage}
    \vspace{-2mm}
    \captionof{figure}{(a) Global attention between coordinates $x_m$ and latents $z_i$ can result in high attention values for non-local latents. (b) Locality is enforced through a Gaussian window $\mu_\sigma(x_m, p_i)$, attenuating the dot-product $\mathbf{q}_{m,i}\mathbf{k}_i$ as a function of the distance between $x_m$ and $p_i$.}
    \label{fig:gaussian-window}
\end{wrapfigure}

To address this issue, we modify the attention mechanism by incorporating a Gaussian spatial windowing function with parameter $\sigma_\text{att}$ into the computation of attention coefficients. This approach follows the strategy proposed by \citet{cordonnier2019relationship}. Specifically, the attention scores derived from the dot product between the query and key values are modulated by a Gaussian window $\mu$, which is dependent on the Euclidean distance between latent positions $p^{pos}$ and the input coordinates, expressed as $\mu_\sigma(x_m, p_i)=-\sigma_\text{att}||p^{pos}_i - x_m||^2$. Here, $\sigma_\text{att}$ is a hyperparameter that regulates the size for each latent (Fig. \ref{fig:gaussian-window}). We have:
\begin{equation}
\label{eq:local_cross_attention}
att(x, z) = \underset{i=1,...,N}{\operatorname{softmax}} \left(\frac{\mathbf{q}(\clr{\mathbf{a}(x, p_i)})^T \mathbf{k}(\mathbf{c}_i)}{\sqrt{d_k}} + \mu_\sigma(\clr{x}, \clr{p_i}) \right),    
\end{equation}
where $\mu(x_m, p_i)$ represents the Gaussian window computed for each latent position. To enhance the expressiveness of the model even further, $\sigma_{att}$ can be made latent-specific, encoding for the spatial extent of a latent $z_i$. This extension allows the latents to be expressed as point clouds: $z^f = \{(p_i, \mathbf{c_i}, \sigma_i)\}^N_{i=1} $, effectively coupling position, appearance, and locality attributes within the latent space. However, we keep this for further research - fixing $\sigma_\text{att}$ in our experiments.

\paragraph{A note on efficiency}
\label{sec:efficient-cross-attention}
A limitation of the proposed method is the considerable computational complexity required to calculate output values for large numbers of input coordinates; larger more complex signals require a larger number of latents to be represented accurately leading to an exponential number of attention coefficients needing to be calculated each forward pass (complexity scales as $O(N_\text{latents}\times N_\text{coordinates})$). Since we localize latents, larger input domains present a trade-off. Either we maintain a small number of latent points—requiring a larger $\sigma_\text{att}$ to cover the entire domain, diminishing locality—or we increase the number of latent points, aggravating computational cost.

To mitigate the computational overhead associated with the cross-attention between the latent points and the sampled coordinates, we propose employing a k-nearest neighbours (k-NN) approach for the cross-attention operation. Specifically, for each pixel, we first identify its k-nearest latent points and then compute cross-attention for this coordinate only over these $k$ nearest latents. This approach reduces computational cost while preserving the advantages of local representations (Appx. \ref{appx:comp-effic}).

\begin{wrapfigure}[17]{r}{0.5\textwidth}
    \centering
    \begin{minipage}[b]{0.32\linewidth}
        \includegraphics[width=\textwidth]{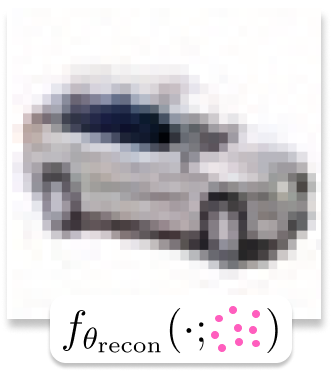}
        \caption*{(a)}
    \end{minipage}
    \hfill
    \begin{minipage}[b]{0.32\linewidth}
        \includegraphics[width=\textwidth]{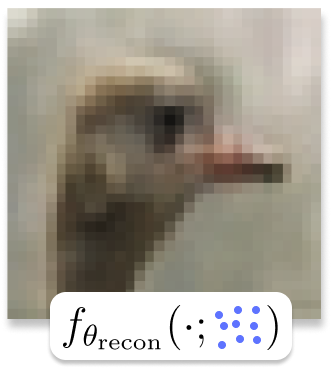}
        \caption*{(b)}
    \end{minipage}
    \hfill
    \begin{minipage}[b]{0.32\linewidth}
        \includegraphics[width=\textwidth]{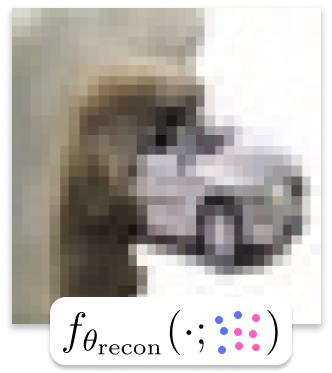}
        \caption*{(c)}
    \end{minipage}
    \caption{Latent point cloud editing. Subfigures (a) and (b) show two reconstructed CIFAR-10 images with corresponding latents $z^\text{car}$,$z^\text{duck}$. Subfigure (c) shows a reconstruction of the latent set $z^\text{car-duck}$ when selecting latents from either $z^\text{car}$ or $z^\text{duck}$ based on their position.}
    \label{fig:carduck}
\end{wrapfigure}

\paragraph{Properties of Equivariant Neural Fields} 
\textit{Weight sharing -} Conditioning ENF's attention operator on attributes $\clr{\mathbf{a}}$ which uniquely identify equivalence classes of (latent-coordinate)-pairs, ensures that the cross-attention operators—be it the attention logits or the value transform—respond similarly regardless of the pose under which a signal pattern presents itself. This form of \textit{weight-sharing} has shown improved data and representation efficiency in GNN-based architectures \citep{bekkers2023fast} and—as such—we hypothesize that our proposed ENF architecture similarly benefits from these properties compared to other types of NeFs. We confirm this property in figure \ref{fig:weight-sharing}, which shows that ENFs share weights over group actions $g \in G$ for the geometry in which the point clouds are grounded. In Sec. \ref{sec:experiments} we verify these benefits on downstream tasks.

\textit{Locality and (geometric) interpretability-} The use of latent point clouds allows for localization of the cross-attention mechanisms \textit{around} the latent pose, akin to how a convolution operator works with localised kernels. Not only does this improve interpretability and downstream performance (field patterns can be attributed to specific latent points), it enables unique field editing possibilities. Since our method is based on sets of latents and each element is responsible for a local neighbourhood of the input domain, we can take arbitrary unions(e.g. stitching) or intersections(e.g. latent-merging) of point-sets of different samples (Fig. \ref{fig:carduck}).


\subsection{Obtaining latent point clouds $z$} \label{subsec:obtaining-z}
Following \cite{park2019deepsdf,dupont2022data} we obtain a latent point cloud $z^f$ for a specific sample $f$ using \textit{gradient descent}, optimizing $z^f$ for reconstruction of the original signal $f$. For instance, in images a latent point cloud $z$ may be optimized for an $L_2$-loss between $f_{\theta}(\cdot,z)$ and $f(\cdot)$. Of course, this also requires optimizing $\theta$ to meaningfully map latents to fields. The two most common approaches to this end are Autodecoding \citep{park2019deepsdf} - where $z^f$ and $\theta$ are optimised simultaneously over a dataset, or MAML \citep{finn2017model,tancik2021learned} - where optimization is split into an outer and inner loop, with $\theta$ being optimized in the outer loop and $z$ being re-initialized every outer step to reconstruct the current signal batch in a limited number of SGD steps in the inner loop. We detail these approaches in Appx. \ref{appx:autodecoding-maml}, using both in the experiments.

\section{Experiments}
\label{sec:experiments}
First, we show the ability of Equivariant Neural Fields (ENFs) to reconstruct datasets of input fields $f$ - i.e. to associate a latent point cloud $z^{f_j}$ to a given dataset of samples ${f_j}\in D$ that accurately reconstructs them. Then, we validate ENFs as an improved NeF-based \textit{downstream representation} for various tasks requiring geometric reasoning; classification, segmentation and forecasting. To show the flexibility of NeF-based representations, we perform these tasks on a range of modalities.

Each downstream experiment consists of two stages: (1) fitting a ENF backbone $f_\theta$ for \textit{reconstruction} of the input fields $f_j\in D$ to obtain latents pointclouds $z^{f_j}$, and (2) training a downstream model—that takes $z^{f_j}$ as input—for each specific task. The bi-invariant $\mathbf{a}_{m,i}$ we choose to condition our ENF, as well as the downstream model, varies depending one the type of task we're performing—described per experiment below. Since our goal is to assess the impact of grounding continuous representations in geometry, besides dataset-specific baselines, we also compare against Functa \citep{dupont2022data}, the framework that originally proposed functional representations $z^{f_j}$ as data surrogates.

For experimental details and hyperparameters we refer to Appx. \ref{appx:experimental-details}.




\subsection{Reconstruction capacity}
\label{sec:reconstruction}

First we evaluate our proposed ENFs on their reconstruction capabilities. \textbullet \; \textbf{Image data} We show results for reconstruction trained with Meta-Learning on CIFAR10 \cite{krizhevsky2009learning}, CelebA$64{\times}64$ \citep{liu2015faceattributes} and ImageNet1K \citep{deng2009imagenet} using different bi-invariant attributes $\mathbf{a}_{m,i}$—resulting in equivariance to different corresponding transformation groups—in Tab. \ref{tab:reconstruction-classification-img}. We provide results for a Functa baseline, following the setup described in \citep{dupont2022data}. Notably, translational weight sharing ($\mathbf{a}^{\mathbb{R}^2}$) outperforms settings with no-transformation ($\mathbf{a}^\emptyset$) and roto-translational weight sharing ($\mathbf{a}^{\rm SE(2)}$). Moreover, it seems that locality alone is itself a useful inductive bias when moving to higher resolution, more varied images; Functa outperforms ENF $\mathbf{a}^\emptyset$ on CIFAR10 reconstruction, but on CelebA and ImageNet all ENF parameterizations outperform Functa. These results reinforce locality and equivariance as inductive biases in image-based continuous reconstruction tasks.
\textbullet \; \textbf{Shape data} We show shape reconstruction results (Tab. \ref{tab:reconstruction-classification-shape}) on two common shape representations; voxels (3D occupancy grids) and meshes. 
For voxel data we take train and test splits from the 16-class ShapeNet-Part segmentation dataset \citep{yi2016scalable} (which we denote ShapeNet16) and fit their corresponding voxel-based representation as occupancy function $\mathbb{R}^3{\rightarrow}\{0,1\}$ - also using the obtained representations in the ShapeNet-Part segmentation experiment detailed below. For mesh data we opt instead to fit the full 55-class ShapeNetCore (v2) object dataset \citep{chang2015shapenet}, fitting these with 150k points sampled from the signed distance function of the mesh (more details in Appx. \ref{sec:shape-datasets-appendix}). Unlike \cite{dupont2022data}, we were unable to get sufficient quality reconstructions with meta-learning and instead obtain latents $z$ using autodecoding on shape data (finding discussed in Appx. \ref{appx:autodecoding-maml}). Results show that ours as well as the baseline models struggle with accurately reconstructing the underlying shape from the SDF point clouds, we think due to the more complex optimization objective.

\begin{table}
    \begin{minipage}{0.49\textwidth}
    \centering
    \caption{Test-set reconstruction PSNR (db$\uparrow$) on CIFAR10, CelebA64x64, ImageNet1k, test accuracy (\%$\uparrow$) on CIFAR10.}
    \label{tab:reconstruction-classification-img}
    \begin{small}
    \scalebox{0.8}{
    \begin{tabular}{lccccc}
    \toprule
     & \multicolumn{2}{c}{\sc{CIFAR10}} & \sc{CelebA} &\sc{ImageNet}\\
     \midrule
     \sc{Task} & \sc{Recon.} & \sc{Class.}& \sc{Recon.}& \sc{Recon.} \\
    \midrule
    Functa & 38.1 &  68.3 & 28.0 & 7.2 \\
    ENF $\mathbf{a}^\emptyset$ &  36.5 & 68.7&  30.6 & 24.7 \\
    ENF $\mathbf{a}^{\mathbb{R}^2}$ & $\boldsymbol{42.2}$ & \textbf{82.1} & $\boldsymbol{34.6}$ & \textbf{27.5} \\
    ENF $\mathbf{a}^{\rm SE(2)}$ & 41.6 & 81.5 & 32.9 & 26.8 \\
    \bottomrule
    \end{tabular}}
    \end{small}
    \end{minipage}%
    \hfill
    \begin{minipage}{0.49\textwidth}
    \centering
    \caption{Test reconstruction (IoU$\uparrow$) on ShapeNet16 and ShapeNet55 and test classification accuracy (\%$\uparrow$) on ShapeNet16.}
    \label{tab:reconstruction-classification-shape}
    \begin{small}
    \scalebox{0.8}{
    \begin{tabular}{lcccc}
    \toprule
     & \multicolumn{2}{c}{\sc{ShapeNet16}}  & \multicolumn{1}{c}{\sc{ShapeNet55}}\\
    \sc{Modality} & \multicolumn{2}{c}{\sc{Voxel (Occ)}} & \multicolumn{1}{c}{\sc{P. Cloud (SDF)}} \\
    \midrule
    \sc{Task} & \sc{Recon.} & \sc{Class.}& \sc{Recon.} \\
    \midrule
    NF2vec  & - & 93.3 & -  \\
    Functa & 92.1 & 90.3 & 25.7  \\
    ENF $\mathbf{a}^{\emptyset}$ & 90.7 & 96.4 & 72.3 \\ 
    ENF $\mathbf{a}^{\mathbb{R}^3}$ & \textbf{92.9} & \textbf{96.6} & \textbf{73.2} \\ 
    \bottomrule
    \end{tabular}}
    \end{small}
    \end{minipage}%
\end{table}

\subsection{Downstream tasks}
\paragraph{Image classification} One major limitation of Functa—noted by \cite{bauer2023spatial}—was lacking performance on complex image tasks such as classification. To show performance of our model in this setting, we reproduce the CIFAR10 classification experiment listed in \cite{bauer2023spatial}—augmenting CIFAR10 with 50 random crops and flips per image, and training an ENF to reconstruct these using meta-learning, obtaining latent point clouds $z$. We do this for different bi-invariants $\mathbf{a}$ corresponding to no equivariance ($\mathbf{a}^\emptyset$), translational ($\mathbf{a}^{\mathbb{R}^2}$) and roto-translational ($\mathbf{a}^{\rm SE(2)}$) equivariance. We then train a P$\Theta$NITA classifier \citep{bekkers2023fast} to classify these latent point clouds - conditioning the message passing function on the same bi-invariants, now calculated between poses $p_i$. Results (Tab. \ref{tab:reconstruction-classification-img}) show a test-accuracy improvement of $13.8$ percentage points ($68.7\%{\rightarrow}82.1\%$) over Functa \citep{dupont2022data}, and also indicate that in this setting (roto-)translational equivariance is a strong inductive bias—with $\mathbf{a}^{\rm SE(2)}$, $\mathbf{a}^{\mathbb{R}^2}$-ENFs outperforming $\mathbf{a}^\emptyset$-ENFs.

\paragraph{Shape classification} Highlighting the flexibility of NeF-representations we apply the same setup to shape classification, training P$\Theta$NITA classifiers on the aforementioned ShapeNet16 dataset. Results in Tab. \ref{tab:reconstruction-classification-shape} show that relevant geometric features are better preserved in a localized latent space. Here, the performance difference between equivariant ($\mathbf{a}^{\mathbb{R}^3}$) and non-equivariant ($\mathbf{a}^\emptyset$) ENFs are negligible. This is to be expected due to the global alignment of the ShapeNet dataset, and shows ENF is able to perform under equivariance constraints even in non-equivariant tasks. 

\begin{wrapfigure}[11]{r}{0.2\textwidth}
    \centering
    \vspace{-4mm}
    \captionof{table}{Mean class and instance IoU ($\uparrow$) on ShapeNet.}
    \vspace{-2mm}
    \label{tab:results_shape_part_short}
    \begin{small}
    \scalebox{0.65}{
        \begin{tabular}{lcc}
        \sc{Model} & \rotatebox[origin=c]{65}{\sc{inst mIoU}} & \rotatebox[origin=c]{65}{\sc{cls mIoU}}\\
        \toprule
        PointNet   & 83.1 & 79.0 \\
        PointNet++ & \textbf{84.9} & \textbf{82.7} \\
        DGCNN      & 83.6 & 80.9 \\
        \midrule
        NF2vec     & 81.3 & \underline{76.9} \\
        Functa     & \underline{82.8} & 74.8 \\
        ENF        & 82.2 & 75.4 \\
        \bottomrule
        \end{tabular}
    }
    \end{small}
    \vspace{-4mm}
\end{wrapfigure}

\paragraph{ShapeNet-Part segmentation}
Where classification primarily evaluates how well the latent point clouds captures global information, we also want to evaluate ENFs performance on fine-grained tasks. As such, we evaluate on the ShapeNet part segmentation task \citep{yi2016scalable}. The ShapeNet-Part dataset consists of point clouds for 16 ShapeNet object-classes, each with a varying number of annotated parts for a total of 50 segmentation classes. We use the ENF $\mathbf{a}^{\mathbb{R}^3}$ backbone defined in the voxel reconstruction task above ($f_{\theta_\text{recon}}$) to obtain a latent $z^\text{recon}$ for a shape. Then, a second ENF ($f_{\theta_\text{seg}}$) is trained to map points on this shape to a one-hot encoding of their corresponding segmentation classes, i.e. $f_{\theta_\text{seg}}(x_m; z^\text{recon})$ maps $x_m$ to its class label $y_m$. Results (Tab. \ref{tab:results_shape_part_short}) somewhat surprisingly show ENF and Functa perform comparably in this task (detailed results and visualizations in Appx. \ref{appx:more-shapenet}). We again think this 
\begin{wrapfigure}[27]{r}{0.5\textwidth}
\vspace{-3mm}
    \centering
    \begin{minipage}{\linewidth}        \includegraphics[width=\linewidth]{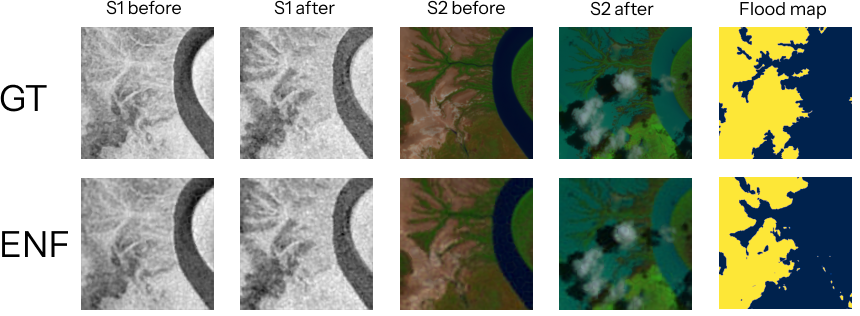}\vspace{-1mm}
    \caption{\textit{Top-} OMBRIA test sample of SAR (S1), optical (S2) before and after flooding with ground truth flood map. \textit{Bottom-} ENF reconstructions $f_{\theta_\text{recon}}(\cdot;z^\text{recon}$) and predicted mask $f_{\theta_\text{seg}}(\cdot;z^\text{recon})$.}
    \label{fig:ombria-seg}
    \end{minipage}
        \begin{minipage}{0.48\linewidth}
        \captionof{table}{Test IOU ($\uparrow$) for flood map segmentation on OMBRIA, for different observation rates.}
        \label{tab:ombria_seg}
        \begin{small}
        \scalebox{0.65}{
            \begin{tabular}{lcc}
            \toprule
            \sc{Model} & PSNR ($\uparrow$) & IoU ($\uparrow$) \\
            \toprule
            & \multicolumn{2}{c}{\sc{100\% of $f_\text{in}$ observed} }\\
            OmbriaNet & N.A. & 72.36 \\
            Functa & 16.77 & 42.75  \\
            ENF &\textbf{31.65} & \textbf{74.00} \\
            \midrule
            & \multicolumn{2}{c}{\sc{50\% of $f_\text{in}$ observed} }\\
            OmbriaNet & N.A. & 27.02 \\
            Functa & 16.71 & 42.74  \\
            ENF & \textbf{31.37} &  \textbf{73.65} \\
            \midrule
            & \multicolumn{2}{c}{\sc{10\% of $f_\text{in}$ observed} }\\
            OmbriaNet & N.A. & 0.0 \\
            Functa & 16.77 & 42.92  \\
            ENF & \textbf{24.87} &  \textbf{71.58} \\
            \bottomrule
            \end{tabular}
        }
        \end{small}
    \end{minipage}%
    \hfill
    \begin{minipage}{0.47\linewidth}
            \centering
            \captionof{table}{Test IOU ($\uparrow$) zero-shot resolution transfer on OMBRIA. $f_{\theta_\text{recon}}$,$f_{\theta_\text{seg}}$ were trained on $128{\times}128$ resolution.}
            \label{tab:ombria_seg_superres}
            \begin{small}
            \scalebox{0.58}{
                \begin{tabular}{lcc}
                \toprule
                \sc{Model} & PSNR ($\uparrow$) & IoU ($\uparrow$) \\
                \toprule
                & \multicolumn{2}{c}{\makecell{\sc{256$\times$256 test resolution}} }\\
                Functa & 16.72 & 37.14  \\
                ENF & \textbf{28.61} & \textbf{72.92}  \\
                \midrule
                & \multicolumn{2}{c}{\makecell{\sc{128$\times$128 test resolution}} }\\
                Functa & 16.71 & 35.48  \\
                ENF & \textbf{29.31} & \textbf{73.21}  \\
                \midrule
                \multicolumn{3}{c}{\makecell{\sc{64$\times$64 test resolution}} }\\
                Functa & 16.58 & 36.90  \\
                ENF & \textbf{33.31} & \textbf{72.50} \\
                \bottomrule
                \end{tabular}
            }
            \end{small}
    \end{minipage}
\end{wrapfigure}
attributable to the fact that all shapes are aligned and centered - we further investigate these results in Appx. \ref{appx:shapenet-addtn}. We additionally include results for point cloud-specific architectures, and NF2Vec - a framework for self-supervised representation learning on 3D data from (non-conditional) NeFs. These results show that ENF only slightly underperforms modality-specific baselines.
\vspace{-3mm}
\paragraph{Flood Map Segmentation}
For a more challenging segmentation task we apply ENFs on multi-modal flood mapping dataset \citep{drakonakis2022ombrianet}. This small dataset (759/85 train/test split) provides dual-modal temporal data; aligned Synthetic Aperture Radar (SAR) and optical satellite images at $256{\times}256$ resolution obtained by satellites Sentinel 1 and 2 (S1,S2), of disaster sites before and after their flooding, along with corresponding masks that segment the flooded area. The goal is to predict binary segmentation mask given these 4 different input fields. We first train a reconstruction $\mathbf{a}^{\rm SE(2)}$-ENF $f_{\theta_\text{recon}}$ with MAML to obtain a latent $z^\text{recon}$ that decodes into the four observations. Next, a segmentation ENF $f_{\theta_\text{seg}}$ is trained to predict, given a latent $z^\text{recon}$, the binary mask at each location.

We evaluate and compare our model against the multi-modal U-Net proposed by \cite{drakonakis2022ombrianet}. As suggested by \cite{dupont2022data} we trained Functa—unable to fit the training set with MAML—using autodecoding instead. However, we found Functa unable to generalize to test images in this complex low-data setting, collapsing to remembering the training dataset (achieving 31.5 recon PSNR and 93.7 IoU on the $256{\times}256$ train set). Results (Tab. \ref{tab:ombria_seg}, Fig. \ref{fig:ombria-seg}) show the importance of inductive biases in complex limited-data regimes. We provide results for \textit{subsampled} observations to simulate missing data, showing the robustness of NeF-based methods to sparsity—ENF performs well even at $10\%$ observation rate where classical convolution-based methods fail. Moreover, we show zero-shot resolution transfer results (Tab. \ref{tab:ombria_seg_superres}) where $f_{\theta_\text{recon}},f_{\theta_\text{seg}}$ are trained on $128{\times}128$ resolution data are deployed on $64\times64$ and $256\times256$ resolution data without fine-tuning, showing the resolution agnostic nature of NeF-representations.

\begin{figure}
    \begin{minipage}{0.32\textwidth}
            \centering
    \captionof{table}{ERA5 reconstruction $T_{t}$-MSE$\downarrow$ and 1-hour forecasting $T_{t+1}$-MSE$\downarrow$. \textit{*MSE between ground truth observations at $T_t$ and $T_{t+1}$.}}
    \label{tab:era5-forecasting}
    \begin{small}
    \scalebox{0.82}{
    \begin{tabular}{lcc}
    \toprule
    & $T_{t}$-\sc{MSE}$\downarrow$ & $T_{t+1}$-\sc{MSE}$\downarrow$  \\
    \toprule
    Identity* & - & 2.42E-05 \\
    Functa & 5.75E-05 & 3.45E-03 \\
    ENF &\textbf{8.04E-06} & \textbf{9.44E-06} \\
    \bottomrule
    \end{tabular}}
    \end{small}
    \end{minipage}
    \hfill
    \begin{minipage}{0.65\textwidth}
    \begin{subfigure}[b]{0.49\textwidth}
        \centering
        \includegraphics[width=\textwidth]{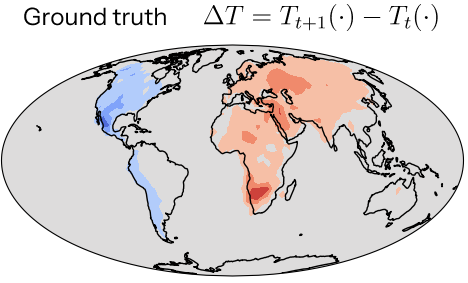}
    \end{subfigure}
    \hfill
    \begin{subfigure}[b]{0.49\textwidth}
        \centering
        \includegraphics[width=\textwidth]{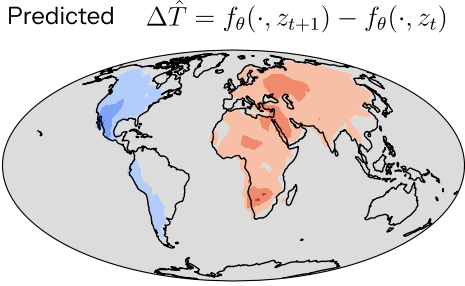}
    \end{subfigure}
    \caption{Visualization of ground truth ERA5 test sample and ENF prediction of the change in temperature between observations $T_{t}$ and $T_{t+1}$. We show the true ($\Delta$T) and predicted ($\Delta\hat{T}$) difference between the temperature maps at $t$ and $t+1$.}
    \vspace{-4mm}
    \label{fig:era5-pred}
    \end{minipage}
\end{figure}

\vspace{-3mm}
\paragraph{ERA5 Climate forecasting} Following \citep{yin2022continuous,knigge2024space} we evaluate our NeF-based representation on dynamics forecasting. ERA5 \citep{hersbach2019era5} is a dataset of hourly global temperature observations. We use the dataset as described in \cite{dupont2021generative}, which contains data defined over $46\times90$ latitude-longitude grids. From the training and test sets, we extract 5693 and 443 pairs of subsequent observations $T_t,T_{t+1}$ for train and test sets respectively. Using MAML, we train an ENF $f_\theta$ with bi-invariant ${\mathbf{a}^\emptyset}$ (no symmetries exist in this data) to reconstruct the global temperature state $T_t$ using $z_t$, and optimize a P$\Theta$NITA MPNN to predict an update $\Delta z_t$ that maps $z_t$ to a latent $\hat{z}_{t+1}=z_t + \Delta z_t$ which decodes into the state at $t+1$, i.e. $T_{t+1}{\approx}f_{\theta}(\cdot; z_{t+1})\approx f_{\theta}(\cdot; z_t + \Delta z_t)$. Training is done sequentially, i.e. first the backbone $f_\theta$ is optimized and afterward the MPNN is trained, keeping $f_\theta$ fixed. Results (Tab. \ref{tab:era5-forecasting}, Fig. \ref{fig:era5-pred}) show that the latent space of ENF lends itself well for modelling such complex dynamics - where a global latent representation such as Functa seems unable to model the relevant fine-grained details needed for forecasting.

\begin{wrapfigure}[13]{r}{0.65\textwidth}
\vspace{-3mm}
\begin{minipage}{0.5\linewidth}
    \includegraphics[width=\linewidth]{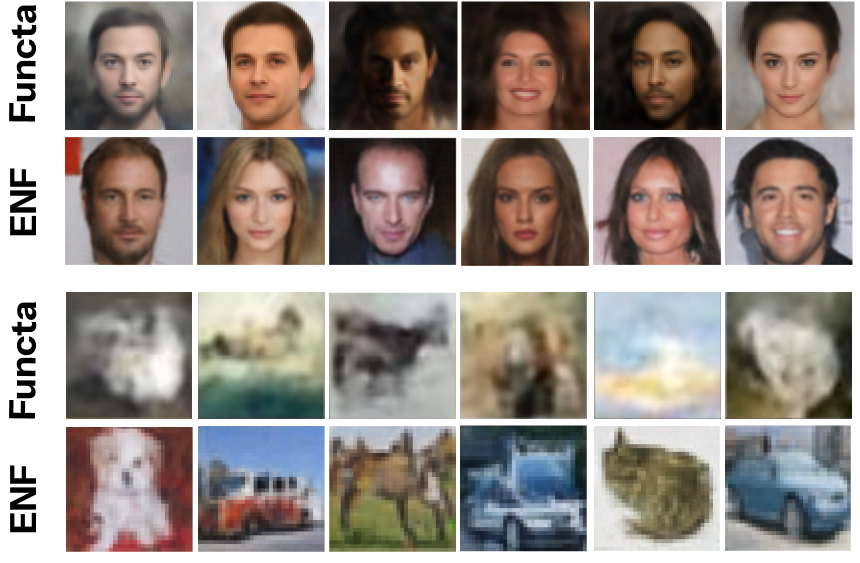}
    \caption{Qualitative samples for generative modelling on CIFAR-10 and Celeba$64{\times}64$.}
    \label{fig:generative}
\end{minipage}\hfill
\begin{minipage}{0.45\linewidth}
    \centering
    \captionof{table}{FID for generative modelling on CIFAR-10 and Celeba$64{\times}64$.}
    \label{tab:generative}
    \begin{small}
    \scalebox{0.70}{
        \begin{tabular}{lcc}
         & CelebA$64{\times}64$  & CIFAR-10 \\
        \sc{Model} & FID $\downarrow$ & FID $\downarrow$ \\
        \toprule
        GEM & - & 23.8 \\
        GASP & 13.5 & - \\
        DPF & 13.2 & 15.1 \\
        \midrule
        Functa     & 40.4 & 78.2 \\
        ENF        & \textbf{33.8 }& \textbf{23.5} \\
        \bottomrule
        \end{tabular}
    }
    \end{small}
\end{minipage}
\end{wrapfigure}

\paragraph{Image generation} Following \cite{dupont2022data, bauer2023spatial}, we provide results for diffusion applied to a dataset of latents obtained from pretrained ENFs on CIFAR10 and Celeba$64\times64$. As downstream diffusion model, we utilize DiT-B \citep{peebles2023scalable}, a natural choice for our set-latent (training detailed in Appx. \ref{appx:generative-modelling}). We provide results in FID \citep{heusel2017gans} for unconditional generation in Tab. \ref{tab:generative} and samples in Fig. \ref{fig:generative}. We provide comparison to Functa \citep{dupont2022data}, as well as other frameworks for generative modelling over fields \citep{du2021learning, dupont2021generative, zhuang2023diffusion}--notably each of these methods is trained on a generative objective and does not support self-supervised pre-training like Functa or ENF. On globally aligned CelebA$64{\times}64$, both Functa and ENF produce perceptually qualitative samples, but unlike ENF, Functa is unable to generalize to CIFAR-10, where data is less homogeneous and not aligned. These results again show clear benefit of a geometrically interpretable latent space for downstream tasks, though previously proposed frameworks specific to generative modelling over fields achieve better performance than ENF. The latter points to a possible area of improvement, and future work could look into incorporating insights from these works into a generative adaptation of the ENF framework, e.g. through latent-space regularization of perceptual consistency as per \cite{du2021learning}.

\section{Conclusion}
\vspace{-3mm}
Building upon fascinating work using Neural Fields (NeFs) as continuous data surrogates, this paper introduces Equivariant Neural Fields (ENFs); a novel NeF parameterization that re-introduces inductive biases (locality, equivariance) into NeF-based representations. ENF uses a geometry-grounded conditioning variable—a latent attributed point cloud—to achieve an equivariant decoding process, ensuring that transformations in an input field are preserved in the latent space and enabling \textit{steering} of the latent to transform the output signal. This steerability property allows for the accurate representation of geometric information, and for efficient weight-sharing over spatially similar patterns, significantly improving learning efficiency and generalization, as validated on a range of experiments with varying data modalities and objectives.  

\section{Reproducibility}
All datasets can be downloaded via their cited references without any effort except for the ShapeNet dataset - which requires registration and approval. The pre-processing steps are described in appendix \ref{sec:shape-datasets-appendix}. For all model parameter settings for the ENFs, downstream models or Functa we refer to the appendix \ref{appx:experimental-details}. As supplementary material we added a codebase containing code to reproduce results for the CIFAR10 and OMBRIA experiments. Code for all other experiments will be released during the rebuttal phase of the review process, containing all settings to reproduce the experiments in config files.

\section{Acknowledgements}
David Knigge is partially funded by Elekta Oncology Systems AB and a RVO public-private partnership grant (PPS2102). David Wessels is partially funded Ellogon.AI and a public grant of the Dutch Cancer Society (KWF) under subsidy (15059/2022-PPS2). This work used the Dutch national e-infrastructure with the support of the SURF Cooperative using grant no. EINF-9549 and EINF-10544.

\bibliography{iclr2025_conference}

\begin{thebibliography}{62}
\providecommand{\natexlab}[1]{#1}
\providecommand{\url}[1]{\texttt{#1}}
\expandafter\ifx\csname urlstyle\endcsname\relax
  \providecommand{\doi}[1]{doi: #1}\else
  \providecommand{\doi}{doi: \begingroup \urlstyle{rm}\Url}\fi

\bibitem[Atzmon et~al.(2022)Atzmon, Nagano, Fidler, Khamis, and Lipman]{atzmon2022frame}
Matan Atzmon, Koki Nagano, Sanja Fidler, Sameh Khamis, and Yaron Lipman.
\newblock Frame averaging for equivariant shape space learning.
\newblock In \emph{Proceedings of the IEEE/CVF Conference on Computer Vision and Pattern Recognition}, pp.\  631--641, 2022.

\bibitem[Batatia et~al.(2022)Batatia, Kovacs, Simm, Ortner, and Cs{\'a}nyi]{batatia2022mace}
Ilyes Batatia, David~P Kovacs, Gregor Simm, Christoph Ortner, and G{\'a}bor Cs{\'a}nyi.
\newblock Mace: Higher order equivariant message passing neural networks for fast and accurate force fields.
\newblock \emph{Advances in Neural Information Processing Systems}, 35:\penalty0 11423--11436, 2022.

\bibitem[Batzner et~al.(2022)Batzner, Musaelian, Sun, Geiger, Mailoa, Kornbluth, Molinari, Smidt, and Kozinsky]{batzner20223}
Simon Batzner, Albert Musaelian, Lixin Sun, Mario Geiger, Jonathan~P Mailoa, Mordechai Kornbluth, Nicola Molinari, Tess~E Smidt, and Boris Kozinsky.
\newblock E (3)-equivariant graph neural networks for data-efficient and accurate interatomic potentials.
\newblock \emph{Nature communications}, 13\penalty0 (1):\penalty0 2453, 2022.

\bibitem[Bauer et~al.(2023)Bauer, Dupont, Brock, Rosenbaum, Schwarz, and Kim]{bauer2023spatial}
Matthias Bauer, Emilien Dupont, Andy Brock, Dan Rosenbaum, Jonathan~Richard Schwarz, and Hyunjik Kim.
\newblock Spatial functa: Scaling functa to imagenet classification and generation.
\newblock \emph{arXiv preprint arXiv:2302.03130}, 2023.

\bibitem[Bekkers(2019)]{bekkers2019b}
Erik~J Bekkers.
\newblock B-spline cnns on lie groups.
\newblock \emph{arXiv preprint arXiv:1909.12057}, 2019.

\bibitem[Bekkers et~al.(2023)Bekkers, Vadgama, Hesselink, van~der Linden, and Romero]{bekkers2023fast}
Erik~J Bekkers, Sharvaree Vadgama, Rob~D Hesselink, Putri~A van~der Linden, and David~W Romero.
\newblock Fast, expressive se $(n) $ equivariant networks through weight-sharing in position-orientation space.
\newblock \emph{arXiv preprint arXiv:2310.02970}, 2023.

\bibitem[Brandstetter et~al.(2021)Brandstetter, Hesselink, van~der Pol, Bekkers, and Welling]{brandstetter2021geometric}
Johannes Brandstetter, Rob Hesselink, Elise van~der Pol, Erik~J Bekkers, and Max Welling.
\newblock Geometric and physical quantities improve e (3) equivariant message passing.
\newblock In \emph{International Conference on Learning Representations}, 2021.

\bibitem[Bronstein et~al.(2021)Bronstein, Bruna, Cohen, and Veli{\v{c}}kovi{\'c}]{bronstein2021geometric}
Michael~M Bronstein, Joan Bruna, Taco Cohen, and Petar Veli{\v{c}}kovi{\'c}.
\newblock Geometric deep learning: Grids, groups, graphs, geodesics, and gauges.
\newblock \emph{arXiv preprint arXiv:2104.13478}, 2021.

\bibitem[Chang et~al.(2015)Chang, Funkhouser, Guibas, Hanrahan, Huang, Li, Savarese, Savva, Song, Su, et~al.]{chang2015shapenet}
Angel~X Chang, Thomas Funkhouser, Leonidas Guibas, Pat Hanrahan, Qixing Huang, Zimo Li, Silvio Savarese, Manolis Savva, Shuran Song, Hao Su, et~al.
\newblock Shapenet: An information-rich 3d model repository.
\newblock \emph{arXiv preprint arXiv:1512.03012}, 2015.

\bibitem[Chatzipantazis et~al.(2022)Chatzipantazis, Pertigkiozoglou, Dobriban, and Daniilidis]{chatzipantazis2022se}
Evangelos Chatzipantazis, Stefanos Pertigkiozoglou, Edgar Dobriban, and Kostas Daniilidis.
\newblock Se (3)-equivariant attention networks for shape reconstruction in function space.
\newblock \emph{arXiv preprint arXiv:2204.02394}, 2022.

\bibitem[Chollet(2017)]{chollet2017xception}
Fran{\c{c}}ois Chollet.
\newblock Xception: Deep learning with depthwise separable convolutions.
\newblock In \emph{Proceedings of the IEEE conference on computer vision and pattern recognition}, pp.\  1251--1258, 2017.

\bibitem[Cohen et~al.(2019)Cohen, Geiger, and Weiler]{cohen2019general}
Taco~S Cohen, Mario Geiger, and Maurice Weiler.
\newblock A general theory of equivariant cnns on homogeneous spaces.
\newblock \emph{Advances in neural information processing systems}, 32, 2019.

\bibitem[Cordonnier et~al.(2019)Cordonnier, Loukas, and Jaggi]{cordonnier2019relationship}
Jean-Baptiste Cordonnier, Andreas Loukas, and Martin Jaggi.
\newblock On the relationship between self-attention and convolutional layers.
\newblock \emph{arXiv preprint arXiv:1911.03584}, 2019.

\bibitem[Corso et~al.(2022)Corso, St{\"a}rk, Jing, Barzilay, and Jaakkola]{corso2022diffdock}
Gabriele Corso, Hannes St{\"a}rk, Bowen Jing, Regina Barzilay, and Tommi~S Jaakkola.
\newblock Diffdock: Diffusion steps, twists, and turns for molecular docking.
\newblock In \emph{The Eleventh International Conference on Learning Representations}, 2022.

\bibitem[Deng et~al.(2021)Deng, Litany, Duan, Poulenard, Tagliasacchi, and Guibas]{deng2021vector}
Congyue Deng, Or~Litany, Yueqi Duan, Adrien Poulenard, Andrea Tagliasacchi, and Leonidas~J Guibas.
\newblock Vector neurons: A general framework for so (3)-equivariant networks.
\newblock In \emph{Proceedings of the IEEE/CVF International Conference on Computer Vision}, pp.\  12200--12209, 2021.

\bibitem[Deng et~al.(2009)Deng, Dong, Socher, Li, Li, and Fei-Fei]{deng2009imagenet}
Jia Deng, Wei Dong, Richard Socher, Li-Jia Li, Kai Li, and Li~Fei-Fei.
\newblock Imagenet: A large-scale hierarchical image database.
\newblock In \emph{2009 IEEE conference on computer vision and pattern recognition}, pp.\  248--255. Ieee, 2009.

\bibitem[Drakonakis et~al.(2022)Drakonakis, Tsagkatakis, Fotiadou, and Tsakalides]{drakonakis2022ombrianet}
Georgios~I Drakonakis, Grigorios Tsagkatakis, Konstantina Fotiadou, and Panagiotis Tsakalides.
\newblock Ombrianet—supervised flood mapping via convolutional neural networks using multitemporal sentinel-1 and sentinel-2 data fusion.
\newblock \emph{IEEE Journal of Selected Topics in Applied Earth Observations and Remote Sensing}, 15:\penalty0 2341--2356, 2022.

\bibitem[Du et~al.(2021)Du, Collins, Tenenbaum, and Sitzmann]{du2021learning}
Yilun Du, Katie Collins, Josh Tenenbaum, and Vincent Sitzmann.
\newblock Learning signal-agnostic manifolds of neural fields.
\newblock \emph{Advances in Neural Information Processing Systems}, 34:\penalty0 8320--8331, 2021.

\bibitem[Dupont et~al.(2021)Dupont, Teh, and Doucet]{dupont2021generative}
Emilien Dupont, Yee~Whye Teh, and Arnaud Doucet.
\newblock Generative models as distributions of functions.
\newblock \emph{arXiv preprint arXiv:2102.04776}, 2021.

\bibitem[Dupont et~al.(2022)Dupont, Kim, Eslami, Rezende, and Rosenbaum]{dupont2022data}
Emilien Dupont, Hyunjik Kim, SM~Eslami, Danilo Rezende, and Dan Rosenbaum.
\newblock From data to functa: Your data point is a function and you can treat it like one.
\newblock \emph{arXiv preprint arXiv:2201.12204}, 2022.

\bibitem[Eijkelboom et~al.(2023)Eijkelboom, Hesselink, and Bekkers]{eijkelboom2023n}
Floor Eijkelboom, Rob Hesselink, and Erik~J Bekkers.
\newblock E $(n) $ equivariant message passing simplicial networks.
\newblock In \emph{International Conference on Machine Learning}, pp.\  9071--9081. PMLR, 2023.

\bibitem[Finn et~al.(2017)Finn, Abbeel, and Levine]{finn2017model}
Chelsea Finn, Pieter Abbeel, and Sergey Levine.
\newblock Model-agnostic meta-learning for fast adaptation of deep networks.
\newblock In \emph{International conference on machine learning}, pp.\  1126--1135. PMLR, 2017.

\bibitem[Gasteiger et~al.(2021)Gasteiger, Becker, and G{\"u}nnemann]{gasteiger2021gemnet}
Johannes Gasteiger, Florian Becker, and Stephan G{\"u}nnemann.
\newblock Gemnet: Universal directional graph neural networks for molecules.
\newblock \emph{Advances in Neural Information Processing Systems}, 34:\penalty0 6790--6802, 2021.

\bibitem[Geirhos et~al.(2018)Geirhos, Rubisch, Michaelis, Bethge, Wichmann, and Brendel]{geirhos2018imagenet}
Robert Geirhos, Patricia Rubisch, Claudio Michaelis, Matthias Bethge, Felix~A Wichmann, and Wieland Brendel.
\newblock Imagenet-trained cnns are biased towards texture; increasing shape bias improves accuracy and robustness.
\newblock \emph{arXiv preprint arXiv:1811.12231}, 2018.

\bibitem[Hermann et~al.(2020)Hermann, Chen, and Kornblith]{hermann2020origins}
Katherine Hermann, Ting Chen, and Simon Kornblith.
\newblock The origins and prevalence of texture bias in convolutional neural networks.
\newblock \emph{Advances in Neural Information Processing Systems}, 33:\penalty0 19000--19015, 2020.

\bibitem[Hersbach et~al.(2019)Hersbach, Bell, Berrisford, Biavati, Hor{\'a}nyi, Mu{\~n}oz~Sabater, Nicolas, Peubey, Radu, Rozum, et~al.]{hersbach2019era5}
Hans Hersbach, Bill Bell, Paul Berrisford, G~Biavati, Andr{\'a}s Hor{\'a}nyi, Joaqu{\'\i}n Mu{\~n}oz~Sabater, Julien Nicolas, C~Peubey, R~Radu, I~Rozum, et~al.
\newblock Era5 monthly averaged data on single levels from 1979 to present.
\newblock \emph{Copernicus Climate Change Service (C3S) Climate Data Store (CDS)}, 10:\penalty0 252--266, 2019.

\bibitem[Heusel et~al.(2017)Heusel, Ramsauer, Unterthiner, Nessler, and Hochreiter]{heusel2017gans}
Martin Heusel, Hubert Ramsauer, Thomas Unterthiner, Bernhard Nessler, and Sepp Hochreiter.
\newblock Gans trained by a two time-scale update rule converge to a local nash equilibrium.
\newblock \emph{Advances in neural information processing systems}, 30, 2017.

\bibitem[Hoogeboom et~al.(2022)Hoogeboom, Satorras, Vignac, and Welling]{hoogeboom2022equivariant}
Emiel Hoogeboom, V{\i}ctor~Garcia Satorras, Cl{\'e}ment Vignac, and Max Welling.
\newblock Equivariant diffusion for molecule generation in 3d.
\newblock In \emph{International conference on machine learning}, pp.\  8867--8887. PMLR, 2022.

\bibitem[Kingma \& Ba(2014)Kingma and Ba]{kingma2014adam}
Diederik~P Kingma and Jimmy Ba.
\newblock Adam: A method for stochastic optimization.
\newblock \emph{arXiv preprint arXiv:1412.6980}, 2014.

\bibitem[Knigge et~al.(2024)Knigge, Wessels, Valperga, Papa, Sonke, Gavves, and Bekkers]{knigge2024space}
David~M Knigge, David~R Wessels, Riccardo Valperga, Samuele Papa, Jan-Jakob Sonke, Efstratios Gavves, and Erik~J Bekkers.
\newblock Space-time continuous pde forecasting using equivariant neural fields.
\newblock \emph{arXiv preprint arXiv:2406.06660}, 2024.

\bibitem[Kofinas et~al.(2024)Kofinas, Knyazev, Zhang, Chen, Burghouts, Gavves, Snoek, and Zhang]{kofinas2024graph}
Miltiadis Kofinas, Boris Knyazev, Yan Zhang, Yunlu Chen, Gertjan~J. Burghouts, Efstratios Gavves, Cees G.~M. Snoek, and David~W. Zhang.
\newblock Graph neural networks for learning equivariant representations of neural networks.
\newblock In \emph{The Twelfth International Conference on Learning Representations}, 2024.
\newblock URL \url{https://openreview.net/forum?id=oO6FsMyDBt}.

\bibitem[Koishekenov \& Bekkers(2023)Koishekenov and Bekkers]{koishekenov2023exploration}
Yeskendir Koishekenov and Erik~J Bekkers.
\newblock An exploration of conditioning methods in graph neural networks.
\newblock \emph{arXiv preprint arXiv:2305.01933}, 2023.

\bibitem[Krizhevsky et~al.(2009)Krizhevsky, Hinton, et~al.]{krizhevsky2009learning}
Alex Krizhevsky, Geoffrey Hinton, et~al.
\newblock Learning multiple layers of features from tiny images.
\newblock 2009.

\bibitem[Liu et~al.(2015)Liu, Luo, Wang, and Tang]{liu2015faceattributes}
Ziwei Liu, Ping Luo, Xiaogang Wang, and Xiaoou Tang.
\newblock Deep learning face attributes in the wild.
\newblock In \emph{Proceedings of International Conference on Computer Vision (ICCV)}, December 2015.

\bibitem[Mescheder et~al.(2019)Mescheder, Oechsle, Niemeyer, Nowozin, and Geiger]{mescheder2019occupancy}
Lars Mescheder, Michael Oechsle, Michael Niemeyer, Sebastian Nowozin, and Andreas Geiger.
\newblock Occupancy networks: Learning 3d reconstruction in function space.
\newblock In \emph{Proceedings of the IEEE/CVF conference on computer vision and pattern recognition}, pp.\  4460--4470, 2019.

\bibitem[Mildenhall et~al.(2021)Mildenhall, Srinivasan, Tancik, Barron, Ramamoorthi, and Ng]{mildenhall2021nerf}
Ben Mildenhall, Pratul~P Srinivasan, Matthew Tancik, Jonathan~T Barron, Ravi Ramamoorthi, and Ren Ng.
\newblock Nerf: Representing scenes as neural radiance fields for view synthesis.
\newblock \emph{Communications of the ACM}, 65\penalty0 (1):\penalty0 99--106, 2021.

\bibitem[Papa et~al.(2023)Papa, Valperga, Knigge, Kofinas, Lippe, Sonke, and Gavves]{papa2023train}
Samuele Papa, Riccardo Valperga, David Knigge, Miltiadis Kofinas, Phillip Lippe, Jan-Jakob Sonke, and Efstratios Gavves.
\newblock How to train neural field representations: A comprehensive study and benchmark.
\newblock \emph{arXiv preprint arXiv:2312.10531}, 2023.

\bibitem[Park et~al.(2019)Park, Florence, Straub, Newcombe, and Lovegrove]{park2019deepsdf}
Jeong~Joon Park, Peter Florence, Julian Straub, Richard Newcombe, and Steven Lovegrove.
\newblock Deepsdf: Learning continuous signed distance functions for shape representation.
\newblock In \emph{Proceedings of the IEEE/CVF conference on computer vision and pattern recognition}, pp.\  165--174, 2019.

\bibitem[Peebles \& Xie(2023)Peebles and Xie]{peebles2023scalable}
William Peebles and Saining Xie.
\newblock Scalable diffusion models with transformers.
\newblock In \emph{Proceedings of the IEEE/CVF International Conference on Computer Vision}, pp.\  4195--4205, 2023.

\bibitem[Perez et~al.(2018)Perez, Strub, De~Vries, Dumoulin, and Courville]{perez2018film}
Ethan Perez, Florian Strub, Harm De~Vries, Vincent Dumoulin, and Aaron Courville.
\newblock Film: Visual reasoning with a general conditioning layer.
\newblock In \emph{Proceedings of the AAAI conference on artificial intelligence}, volume~32, 2018.

\bibitem[Rahaman et~al.(2019)Rahaman, Baratin, Arpit, Draxler, Lin, Hamprecht, Bengio, and Courville]{rahaman2019spectral}
Nasim Rahaman, Aristide Baratin, Devansh Arpit, Felix Draxler, Min Lin, Fred Hamprecht, Yoshua Bengio, and Aaron Courville.
\newblock On the spectral bias of neural networks.
\newblock In \emph{International conference on machine learning}, pp.\  5301--5310. PMLR, 2019.

\bibitem[Salimans \& Ho(2022)Salimans and Ho]{salimans2022progressive}
Tim Salimans and Jonathan Ho.
\newblock Progressive distillation for fast sampling of diffusion models.
\newblock \emph{arXiv preprint arXiv:2202.00512}, 2022.

\bibitem[Satorras et~al.(2021)Satorras, Hoogeboom, and Welling]{satorras2021n}
V{\i}ctor~Garcia Satorras, Emiel Hoogeboom, and Max Welling.
\newblock E (n) equivariant graph neural networks.
\newblock In \emph{International conference on machine learning}, pp.\  9323--9332. PMLR, 2021.

\bibitem[Sitzmann et~al.(2020{\natexlab{a}})Sitzmann, Chan, Tucker, Snavely, and Wetzstein]{sitzmann2020metasdf}
Vincent Sitzmann, Eric Chan, Richard Tucker, Noah Snavely, and Gordon Wetzstein.
\newblock Metasdf: Meta-learning signed distance functions.
\newblock \emph{Advances in Neural Information Processing Systems}, 33:\penalty0 10136--10147, 2020{\natexlab{a}}.

\bibitem[Sitzmann et~al.(2020{\natexlab{b}})Sitzmann, Martel, Bergman, Lindell, and Wetzstein]{sitzmann2020implicit}
Vincent Sitzmann, Julien Martel, Alexander Bergman, David Lindell, and Gordon Wetzstein.
\newblock Implicit neural representations with periodic activation functions.
\newblock \emph{Advances in neural information processing systems}, 33:\penalty0 7462--7473, 2020{\natexlab{b}}.

\bibitem[Song et~al.(2020)Song, Meng, and Ermon]{song2020denoising}
Jiaming Song, Chenlin Meng, and Stefano Ermon.
\newblock Denoising diffusion implicit models.
\newblock \emph{arXiv preprint arXiv:2010.02502}, 2020.

\bibitem[Tancik et~al.(2020)Tancik, Srinivasan, Mildenhall, Fridovich-Keil, Raghavan, Singhal, Ramamoorthi, Barron, and Ng]{tancik2020fourier}
Matthew Tancik, Pratul Srinivasan, Ben Mildenhall, Sara Fridovich-Keil, Nithin Raghavan, Utkarsh Singhal, Ravi Ramamoorthi, Jonathan Barron, and Ren Ng.
\newblock Fourier features let networks learn high frequency functions in low dimensional domains.
\newblock \emph{Advances in neural information processing systems}, 33:\penalty0 7537--7547, 2020.

\bibitem[Tancik et~al.(2021)Tancik, Mildenhall, Wang, Schmidt, Srinivasan, Barron, and Ng]{tancik2021learned}
Matthew Tancik, Ben Mildenhall, Terrance Wang, Divi Schmidt, Pratul~P Srinivasan, Jonathan~T Barron, and Ren Ng.
\newblock Learned initializations for optimizing coordinate-based neural representations.
\newblock In \emph{Proceedings of the IEEE/CVF Conference on Computer Vision and Pattern Recognition}, pp.\  2846--2855, 2021.

\bibitem[Thomas et~al.(2018)Thomas, Smidt, Kearnes, Yang, Li, Kohlhoff, and Riley]{thomas2018tensor}
Nathaniel Thomas, Tess Smidt, Steven Kearnes, Lusann Yang, Li~Li, Kai Kohlhoff, and Patrick Riley.
\newblock Tensor field networks: Rotation-and translation-equivariant neural networks for 3d point clouds.
\newblock \emph{arXiv preprint arXiv:1802.08219}, 2018.

\bibitem[Vadgama et~al.(2022)Vadgama, Tomczak, and Bekkers]{vadgama2022kendall}
Sharvaree Vadgama, Jakub~Mikolaj Tomczak, and Erik~J Bekkers.
\newblock Kendall shape-vae: Learning shapes in a generative framework.
\newblock In \emph{NeurIPS 2022 Workshop on Symmetry and Geometry in Neural Representations}, 2022.

\bibitem[Vadgama et~al.(2023)Vadgama, Tomczak, and Bekkers]{vadgama2023continuous}
Sharvaree Vadgama, Jakub~M Tomczak, and Erik Bekkers.
\newblock Continuous kendall shape variational autoencoders.
\newblock In \emph{International Conference on Geometric Science of Information}, pp.\  73--81. Springer, 2023.

\bibitem[Van~Quang et~al.(2019)Van~Quang, Chun, and Tokuyama]{van2019capsulenet}
Nguyen Van~Quang, Jinhee Chun, and Takeshi Tokuyama.
\newblock Capsulenet for micro-expression recognition.
\newblock In \emph{2019 14th IEEE International Conference on Automatic Face \& Gesture Recognition (FG 2019)}, pp.\  1--7. IEEE, 2019.

\bibitem[Weiler \& Cesa(2019)Weiler and Cesa]{weiler2019general}
Maurice Weiler and Gabriele Cesa.
\newblock General e (2)-equivariant steerable cnns.
\newblock \emph{Advances in neural information processing systems}, 32, 2019.

\bibitem[Williams(2022)]{point-cloud-utils}
Francis Williams.
\newblock Point cloud utils, 2022.
\newblock https://www.github.com/fwilliams/point-cloud-utils.

\bibitem[Xie et~al.(2022)Xie, Takikawa, Saito, Litany, Yan, Khan, Tombari, Tompkin, Sitzmann, and Sridhar]{xie2022neural}
Yiheng Xie, Towaki Takikawa, Shunsuke Saito, Or~Litany, Shiqin Yan, Numair Khan, Federico Tombari, James Tompkin, Vincent Sitzmann, and Srinath Sridhar.
\newblock Neural fields in visual computing and beyond.
\newblock In \emph{Computer Graphics Forum}, volume~41, pp.\  641--676. Wiley Online Library, 2022.

\bibitem[Yi et~al.(2016)Yi, Kim, Ceylan, Shen, Yan, Su, Lu, Huang, Sheffer, and Guibas]{yi2016scalable}
Li~Yi, Vladimir~G Kim, Duygu Ceylan, I-Chao Shen, Mengyan Yan, Hao Su, Cewu Lu, Qixing Huang, Alla Sheffer, and Leonidas Guibas.
\newblock A scalable active framework for region annotation in 3d shape collections.
\newblock \emph{ACM Transactions on Graphics (ToG)}, 35\penalty0 (6):\penalty0 1--12, 2016.

\bibitem[Yim et~al.(2023)Yim, Campbell, Foong, Gastegger, Jim{\'e}nez-Luna, Lewis, Satorras, Veeling, Barzilay, Jaakkola, et~al.]{yim2023fast}
Jason Yim, Andrew Campbell, Andrew~YK Foong, Michael Gastegger, Jos{\'e} Jim{\'e}nez-Luna, Sarah Lewis, Victor~Garcia Satorras, Bastiaan~S Veeling, Regina Barzilay, Tommi Jaakkola, et~al.
\newblock Fast protein backbone generation with se (3) flow matching.
\newblock \emph{arXiv preprint arXiv:2310.05297}, 2023.

\bibitem[Yin et~al.(2022)Yin, Kirchmeyer, Franceschi, Rakotomamonjy, and Gallinari]{yin2022continuous}
Yuan Yin, Matthieu Kirchmeyer, Jean-Yves Franceschi, Alain Rakotomamonjy, and Patrick Gallinari.
\newblock Continuous pde dynamics forecasting with implicit neural representations.
\newblock \emph{arXiv preprint arXiv:2209.14855}, 2022.

\bibitem[Zhang et~al.(2023)Zhang, Tang, Niessner, and Wonka]{zhang20233dshape2vecset}
Biao Zhang, Jiapeng Tang, Matthias Niessner, and Peter Wonka.
\newblock 3dshape2vecset: A 3d shape representation for neural fields and generative diffusion models.
\newblock \emph{ACM Transactions on Graphics (TOG)}, 42\penalty0 (4):\penalty0 1--16, 2023.

\bibitem[Zhao et~al.(2021)Zhao, Jiang, Jia, Torr, and Koltun]{zhao2021point}
Hengshuang Zhao, Li~Jiang, Jiaya Jia, Philip~HS Torr, and Vladlen Koltun.
\newblock Point transformer.
\newblock In \emph{Proceedings of the IEEE/CVF international conference on computer vision}, pp.\  16259--16268, 2021.

\bibitem[Zhuang et~al.(2023)Zhuang, Abnar, Gu, Schwing, Susskind, and Bautista]{zhuang2023diffusion}
Peiye Zhuang, Samira Abnar, Jiatao Gu, Alex Schwing, Joshua~M Susskind, and Miguel~Angel Bautista.
\newblock Diffusion probabilistic fields.
\newblock In \emph{The Eleventh International Conference on Learning Representations}, 2023.

\bibitem[Zintgraf et~al.(2019)Zintgraf, Shiarli, Kurin, Hofmann, and Whiteson]{zintgraf2019fast}
Luisa Zintgraf, Kyriacos Shiarli, Vitaly Kurin, Katja Hofmann, and Shimon Whiteson.
\newblock Fast context adaptation via meta-learning.
\newblock In \emph{International Conference on Machine Learning}, pp.\  7693--7702. PMLR, 2019.

\end{thebibliography}
\bibliographystyle{iclr2025_conference}

\appendix
\section{Appendix}
\subsection{Autodecoding and Meta-Learning}
\subsubsection{Meta-learning}
\label{appx:autodecoding-maml}
When fitting samples with (Conditional) NeFs using autodecoding (gradient-descent based optimisation of the latent at test-time) \citep{park2019deepsdf}, two key challenges emerge: (1) optimising the sample-specific parameters/latent $z$ for a novel sample can be time-consuming -taking e.g. up to 500 gradient updates \citep{yin2022continuous}- and (2) more gradient updates to NeF weights may impede downstream performance through a phenomenon known as \textit{overtraining} \citep{papa2023train} -where the relationship between field $f$ and $z$ is obscured by oversensitivity to high-frequency details. To address the first point, \cite{sitzmann2020metasdf, tancik2021learned, dupont2022data} propose a Model-Agnostic Meta-Learning (MAML) \citep{finn2017model} based optimisation method, enabling the network or latent initialisation to be learned such that each sample can be fitted with just a few gradient steps. More specifically, \cite{dupont2022data} proposes an inner-outer loop structure where the modulations are updated in the inner loop, while the base network weights are updated in the outer loop. This method corresponds to an instance of learning a subset of weights with MAML, also known as Contextual Variable Interaction Analysis (CAVIA) \cite{zintgraf2019fast}.  More recently, \cite{knigge2024space} note that this meta-learning approach also improves downstream performance by imposing structure on the NeF's latent-space. We provide pseudocode for this approach in Alg. \ref{alg:meta-learning}.

\begin{algorithm}[!htb]
\caption{Meta-learning ENF}\label{alg:meta-learning}
\begin{algorithmic}
\item Randomly initialize shared base network $f_\theta$
\While {not done} 

\indent Sample batch of signals $f$ 

\indent Sample random coordinates $\mathbf{x}$

\indent Initialize latents $z^f \leftarrow \{(p_i, \mathbf{c}_i)\}_{i=1}^N$ for a batch of signals.

\ForAll {step $\in {1, ..., N_{\text{inner}}}$ and $j \in \mathcal{B}$}

\indent \indent $z^f \leftarrow z^f - \epsilon \nabla_{z^f} \mathcal{L}_\text{mse}\big(f_{\theta}(x, z^f), f(x))\big)$
\EndFor

\item Update ENF: $\theta \leftarrow \theta - \eta \nabla_\theta \mathcal{L}_\text{mse}'$
\EndWhile
\end{algorithmic}
\end{algorithm}

\subsubsection{Autodecoding}
During our experiments, we found that not all types of signals lend themselves easily to this encoding approach when using ENFs (specifically SDFs and occupancy functions). Although it saves time in inference and adds structure to the latent space, \citep{dupont2022data} also remark on the limited expressivity of Meta-Learning due to the small number of gradient descent steps used to optimize a latent $z$. As such, for all shape experiments we instead opt for autodecoding \citep{park2019deepsdf}, in which latents and backbone are optimized simultaneously. We provide pseudocode for this approach in Alg. \ref{alg:autodecoding}.

\begin{algorithm}[!htb]
\caption{Autodecoding ENF}\label{alg:autodecoding}
\begin{algorithmic}
\item Randomly initialize shared base network $f_\theta$
\item Initialize latents $z^f_0 \leftarrow \{(p_i, \mathbf{c}_i)\}_{i=1}^N$ \textbf{for all signals}
\While {not done} 

\indent Sample batch of signals $f$ 

\indent Sample random coordinates $\mathbf{x}$

\indent Update latent: $z_{t+1}^f \leftarrow z_{t}^f - \epsilon \nabla_{z_{t}^f} \mathcal{L}_\text{mse}\big(f_{\theta}(x, z_{t}^f), f(x))\big)$

\indent Update ENF: $\theta_{t+1} \leftarrow \theta_t - \eta \nabla_{\theta_t} \mathcal{L}_\text{mse}\big(f_{\theta_t}(x, z_{t}^f), f(x))\big)$
\EndWhile
\end{algorithmic}
\end{algorithm}

\subsubsection{A note on pose initialization} We noted during our experiments that initialization of the latent poses-i.e. their initial position/orientation in the inner loop-has a significant impact on the reconstruction capacity and stability of the ENF. We found that a good way to initialize the latents is to space them as equidistantly as possible and then adding small Gaussian noise ($~N(0, 1e-3)$), e.g. for 2D images on a perturbed 2D grid. Any orientations are initialized canonically, i.e. all latents are initialized with the same orientation. When defining an equidistantly spaced grid is hard, for example on point clouds or data defined on a sphere, we propose using Farthest Point Sampling on a training grid to initialize positions for the latents. 


\section{Bi-invariant function parameterizations $\mathbf{a}_{m,i}$}
\label{appx:inv_attributes}
The bi-invariants attributes that are used in the experiments section are listed here.

\textit{Translational symmetries $\mathbb{R}^n$} In this setting, poses correspond to translations $\mathbf{t}_i\in\mathbb{R}^n$:
\begin{equation}
\label{eq:equiv-translation}
    \mathbf{a}^{\mathbb{R}^n}_{m,i} = x_m - \mathbf{t}_i
\end{equation}

\textit{Roto-translational symmetries ${\rm SE(2)}$}. In this setting, poses $p_i$ correspond to group elements $g=(\theta_i, \mathbf{t}_i)\in {\rm SE(2)}$. We adopt the invariant attribute introduced by \citep{bekkers2023fast}:
\begin{equation}
\label{eq:rototrans}
    \mathbf{a}^{\rm SE(2)}_{m,i} = \mathbf{R}_{\theta_i}(x_m - \mathbf{t}_i)
\end{equation}

\textit{No transformation symmetries}. A simple "bi-invariant" for this setting that preserves all geometric information is given by simply concatenating coordinates $p$ with coordinates $x$: 
\begin{equation}
\label{eq:equiv-class-nosymm}
    \mathbf{a}^{\emptyset}_{i,m} = p_i \oplus x_m
\end{equation}
Parameterizing the cross-attention operation in Eq. \ref{eq:cross-attn} as function of this bi-invariant results in a framework without any equivariance constraints. We use this in experiments to ablate over equivariance constraints and its impact on performance.

\section{Experimental details}
\label{appx:experimental-details}

We provide hyperparameters per experiment. We optimize the weights of the neural field $f_\theta$ in all experiments with Adam \citep{kingma2014adam} with a learning rate of 1e-4, and an inner step size of 30.0 for $\mathbf{c}_i$ and 1.0 for $p_i$ (increasing inner step size in general speeds up convergence and improves reconstruction - but may also lead to instabilities). For downstream classification we train an equivariant MPNN $F_\psi$, using 3 message passing layers in the architecture specified in \cite{bekkers2023fast} conditioned on the same bi-invariant which was used to fit the ENF, with a hidden dimensionality of 256, always trained with learning rate 1e-4. The std parameters $\sigma_q,\sigma_v$ of the RFF embedding functions $\varphi_q,\varphi_v$ are chosen per experiment based on an ablation. In general, increasing both values leads to increased frequency response of the ENF, though generally the model is more susceptible to small change in $\sigma_q$. as well as hidden dim size and the number of attention heads are chosen per experiment, detailed below. We run all experiments on a single H100.

\subsection{Image reconstruction, classification, segmentation, forecasting}

\paragraph{CIFAR10 reconstruction and classification} For CIFAR10 \citep{krizhevsky2009learning} reconstruction and classification we use a hidden dim of 128 with 3 heads, 25 latents of size 64, a batch size of 32 and restrict the cross-attention operator to k=4 nearest latents for each input coordinate $x$. For $\sigma_q,\sigma_v$ we choose 1.0 and 3.0 respectively. We train the ENF model and the classifier for 100 epochs.

\paragraph{CelebA{64$\times$64}} For CelebA \cite{liu2015faceattributes} we use a hidden dim of 256, 36 latents of size 64, a batch size of 2 and restrict the cross-attention operator to k=4 nearest latents for each input coordinate $x$. For $\sigma_q,\sigma_v$ we choose 2.0 and 10.0. We train the model for 30 epochs.

\paragraph{ImageNet1K reconstruction} For ImageNet1K \citep{deng2009imagenet} reconstruction we use a hidden dim of 128 with 3 heads, 169 latents of size 64, a batch size of 2, restricting the cross-attention operator to k=4 nearest latent for each input coordinate $x$. For $\sigma_q,\sigma_v$ we choose 2.0 and 10.0 respectively. We train the model for 2 epochs.

\paragraph{Ombria} For OMBRIA \cite{drakonakis2022ombrianet} we trained a reconstruction model $f_{\theta_\text{recon}}$ with, 256 hidden dim, 4 heads, 169 latents of size 128, a batch size 8. Restricting the cross-attention operator to k=1 nearest latent for each input coordinate $x$. For $\sigma_q,\sigma_v$ we choose 2.0 and 10.0 respectively. The segmentation model $f_{\theta_\text{seg}}$ has hidden size of 128, 8 heads, with cross-attention restricted to k=4 nearest latents, trained with batch size 16. For $\sigma_q,\sigma_v$ we choose 2.0 and 3.0 respectively. We train both models, sequentially, for 500 epochs.

\paragraph{ERA5 forecasting} For ERA5 forecasting \citep{hersbach2019era5} we train a reconstruction model $f_{\theta_\text{recon}}$ with 128 hidden dim, 3 heads, 36 latents of size 64, a batch size of 32. Restricting the cross-attention operator to k=4 nearest latent for each input coordinate $x$. For $\sigma_q,\sigma_v$ we choose 2.0 and 8.0 respectively. Inputs are defined over a latitude longitude $\theta,\phi$ grid, which we map to 3D euclidean coordinates per $\mathbf{x}=\left[\cos\theta \cos \phi, \cos\theta\sin\phi, \sin\theta\right]$. We first train the ENF for 800 epochs. As forecasting model, we train a P$\Theta$NITA MPNN of 3 layers with 256 hidden dim for 1000 epochs. Both models are trained with a batch size of 32.

As objective, since we don't want to overfit the reconstruction error incurred by fitting a latent $z_t$ to the initial state, we supervise the forecasting model with $L_2$ loss between the decoded output for predicted latent $\hat{z}_{t+1}$ per: 
$$ L_\text{forecast}=||(f_{\theta_\text{recon}}(\cdot; z_t) + \Delta T) - (f_{\theta_\text{recon}}(\cdot; \hat{z}_{t+1})||_2^2$$
$\Delta T$ being the ground truth change in temperature, and $\hat{z}_{t+1}=F_{\psi_\text{forecast}}(z_t)$.

\subsection{Functa baseline models}
For the Functa baselines \citep{dupont2022data}, we try to keep as close as possible to the setup defined by the original authors. However, we found that training deeper models in the shape experiments ($\geq$ 8 layers) lead to very unstable training. Instead, for these experiments, we opted to go for shallower models, up to 6 layers. For all experiments we use a hidden dim of 512 and latent modulation size of 512 as used in \citep{bauer2023spatial}, except for ImageNet1K reconstruction, where we use a 1024 latent modulation. We would like to note here that in all experiments, the Functa baseline has larger parameter count than the ENF models applied to each task (e.g. for CIFAR10, ENF has 522K params where Functa has 2.6M params). Although we did not explore this in-depth, it seems that the proposed ENF representation is more parameter efficient compared to the deep SIREN model defined in \citep{dupont2022data}.

For downstream models we follow \citep{bauer2023spatial} and use a 1024 hidden dim 3 layer residual MLP ($\sim$2.1M params). Like in our ENF experiments, we use the same architecture across tasks, only changing the output head to accommodate.

\subsection{Shape reconstruction and classification} 
\paragraph{Voxel Dataset and Segmentation}
\label{sec:shape-datasets-appendix}
The voxels are given with the ShapeNet dataset where the segmentation labels are given as point clouds. However, the coordinate frames of the voxels are different, so to align them we mapped both between -1 and 1.

We trained a model $f_{\theta_\text{recon}}$ with a hidden dim of 128, 3 heads, 27 latents of size 32.  We set $\sigma_q, \sigma_v$ for the RFF embedding functions $\phi_q, \phi_v$ to 2 and 10 respectively. 
For Functa we used, a latent dim of 864 to have the same latent parameters as the conditioning variable used for ENF. As NeF we used a 5-layer Siren with a hidden-dim of 512, $w_0$ is set to 10. The modulation network is a two-layer MLP of hidden sizes 256 and 512.

As is customary for ShapeNet-part segmentation, we condition on the object class and supervise over all segmentation classes with a cross-entropy loss, but only calculate test IoU based on segmentation classes that correspond to the object class. We chose the class-emb dim to be 32 for all settings. 

The segmentation NeFs are all trained for 500 epochs with the same parameters as the reconstruction NeF. However, for ENF, we chose $\sigma_q, \sigma_v$ to be 1, 1 for extra stability.

\paragraph{SDF Dataset}
To create the signed distance functions from ShapeNetCore V2 \cite{chang2015shapenet} objects, we took their meshes and made them water-tight using Point Cloud Utils \citep{point-cloud-utils}. Afterwards, we sampled a point cloud of 150,000 points from the surface. To create the actual SDF, we perturbed the points with Gaussian noise along the mesh normals, and recalculate the signed distances to the surface. Finally, the dataset consisted of 55 classes with a total number of 42.472 and 5.000 samples for the train and test set respectively. 

For ENF we used a latent point cloud of 27 points with context vectors of dimension 32. The std parameters $\sigma_q, \sigma_v$ for the RFF embedding functions $\phi_q, \phi_v$ are 2 and 10 respectively. The hidden dim of the ENF was set to 128 and we used 3 attention-heads. The 

For Functa \citep{dupont2022data} we used a latent modulation of 864 which corresponds to ENFs chosen 27*32 parameters for the conditioning variable. As a NeF we used a 5 layer Siren with an hidden-dim of 512 and a $w_0$ parameter of 15. As a modulation network we used a two-layer MLP with 256,512 hidden-dim.

\subsection{Generative modelling on ENF latent space}
\label{appx:generative-modelling}
For generative modelling experiments CelebA$64{\times}64$ and CIFAR-10 we train a Diffusion Transformer (DiT-B) \citep{peebles2023scalable} on ENF latents, utilizing the context vectors as input tokens and their positions as input for an RFF position embedding added to the tokens. We use the same approach for CelebA as for CIFAR10; we train an $\mathbf{a}^{\mathbb{R}^2}$ ENF with MAML on the image dataset, and use this model to obtain sets of "ground truth" latents $z_0:=\{p_{i}, \mathbf{c}_{i,0} \}_{i=1}^N$ for each image. We then train a DiT-B on a diffusion objective on this latent space, where the forward diffusion kernel is given by: 
\begin{equation}
    z_t = \{(p_{i}, \sqrt{\bar{\alpha}_t}\mathbf{c}_{i, 0} + \sqrt{1-\bar{\alpha}_t}\epsilon^\mathbf{c})\}_{i=1}^N,
\end{equation}
with $\epsilon^\mathbf{c} \in N(0, 1)$, i.e. we only add noise to the latent vectors, as we find adding noise to the poses leads to unstable training (something to be investigated in future work). To generate a sample, we take a random set of "ground truth" poses from the training set, and attach a context vector $\mathbf{c}_{i, t}\in N(0, 1)$ to each pose to denoise. We supervise the DiT-B with the v objective \citep{salimans2022progressive}. Like \citep{peebles2023scalable}, we use a $t_\text{max}{=}1000$ linear noise schedule ranging from $1e{-}4$ to $1e{-}2$, and generate samples using DDIM \citep{song2020denoising} in 512 steps.

In both settings, we train the diffusion model for 200 epochs using Adam, a constant learning rate of $1e{-}4$, no weight-decay or dropout, and generate 50k samples to calculate FID.
\color{black}
\section{Additional results}

\subsection{Size of latent point-cloud}
In this section, we delve deeper into the hyper-parameters of the latent point clouds used as conditioning variables. Equivariant Neural Fields can increase the number of parameters used to represent a signal in two ways: by increasing the number of latent points or by expanding the dimensionality of the context vectors. Intuitively, we can either enhance the representational capability of a single region in the input domain or create more, smaller regions with lower representational dimensions.

To gain further insights into how these conditioning variables behave, we train multiple CNFs using different latent point-cloud configurations. In these experiments, the chosen latent dimension or number of latent points is adjusted to keep the total number of parameters as close as possible across configurations. We trained each model for 400 epochs, as only small improvements occurred beyond this point and the overall trend was already clear. After fitting the ENF, we used the meta-learned representation to perform classification tasks. We employed a simple message-passing GNN, which we trained for 20 epochs, after which performance improvements began to degrade. Below, we present the ablation results for these different approaches to increasing representational capabilities.

\begin{table}[!htb]
    \begin{minipage}{\textwidth}
    \centering
    \caption{Reconstruction PSNR (db$\uparrow$) and ACC (\%$\uparrow$) on CIFAR10 for different parametrisations of the latent point-clouds, i.e. varying $N,d$ in $z:=\{\mathbf{c}_i \in \mathbb{R}^d\}_{i=1}^N$.}
    \label{tab:pointcloud_abl}
    \begin{small}
    \scalebox{0.8}{
    \begin{tabular}{ccccc}
    \toprule
    \sc{\# Latents}(N) & \sc{Latent Dim}(d) & \sc{\# params} & \sc{PSNR} & \sc{ACC (\%)} \\
    \toprule
    1 & 1600 & 1600 & 22.69 & 53.21 \\
    4 & 400 & 1600 & 29.14 & 64.98 \\
    9 & 178 & 1602 & 35.49 & 73.54 \\ 
    16 & 100 & 1600 & 39.93 & 77.09 \\
    \bottomrule
    \end{tabular}}
    \end{small}
    \end{minipage}%
\end{table} 

\subsection{Ablation on Gaussian spatial windowing and kNN approximation}
To evaluate the impact of Gaussian spatial windowing (GSW) and the k-Nearest Neighbors (kNN) approximation in the proposed method, we trained four models on the CIFAR10 dataset: one with both features disabled, one with only kNN enabled, one with only GSW enabled, and one with both enabled. Besides evaluating the difference in reconstruction capabilities, we are mainly interested in the downstream performance. We argue that the introduced locality enhances latent-space structure by improving weight-sharing across local-patches.

After training the models, we used the different ENF models to generate latent representations for CIFAR10 classification. The results are shown in Table \ref{tab:gws-knn-ablation}. While reconstruction performance remains almost consistent across the different setups, GWS significantly improves downstream classification accuracy. Moreover, the kNN approximation does not negatively affect either reconstruction nor classification performance. Interestingly, kNN even provides a slight improvement even without GWS. We hypothesize that this improvement comes from kNN introducing an implicit form of windowing—not by modifying attention values directly but by limiting the set of attention values considered. To conclude, there can be observed that the introduced locality in CNF latents does improve the downstream performance. 

\begin{table}[!htb]
    \begin{minipage}{\textwidth}
    \centering
    \caption{Reconstruction PSNR (db$\uparrow$) and ACC (\%$\uparrow$) on CIFAR10 to ablate the Gaussian spatial windowing and kNN approximation.}
    \label{tab:gws-knn-ablation}
    \begin{small}
    \scalebox{0.8}{
    \begin{tabular}{lcc}
    \toprule
    \sc{ENF setups} & \sc{PSNR (db $\uparrow$)} & \sc{ACC (\% $\uparrow$)} \\
    \toprule
    ENF & 39.1 & 70.1 \\
    ENF + kNN & 40.8 & 72.8 \\
    ENF + GWS & TBD & TBD \\
    ENF + GWS + kNN & 42.2 & 82.1 \\
    \bottomrule
    \end{tabular}}
    \end{small}
    \end{minipage}%
\end{table} 

\subsection{Transforming the latent point-cloud}
\begin{figure}[!htb]
    \centering
    \includegraphics[scale=0.2]{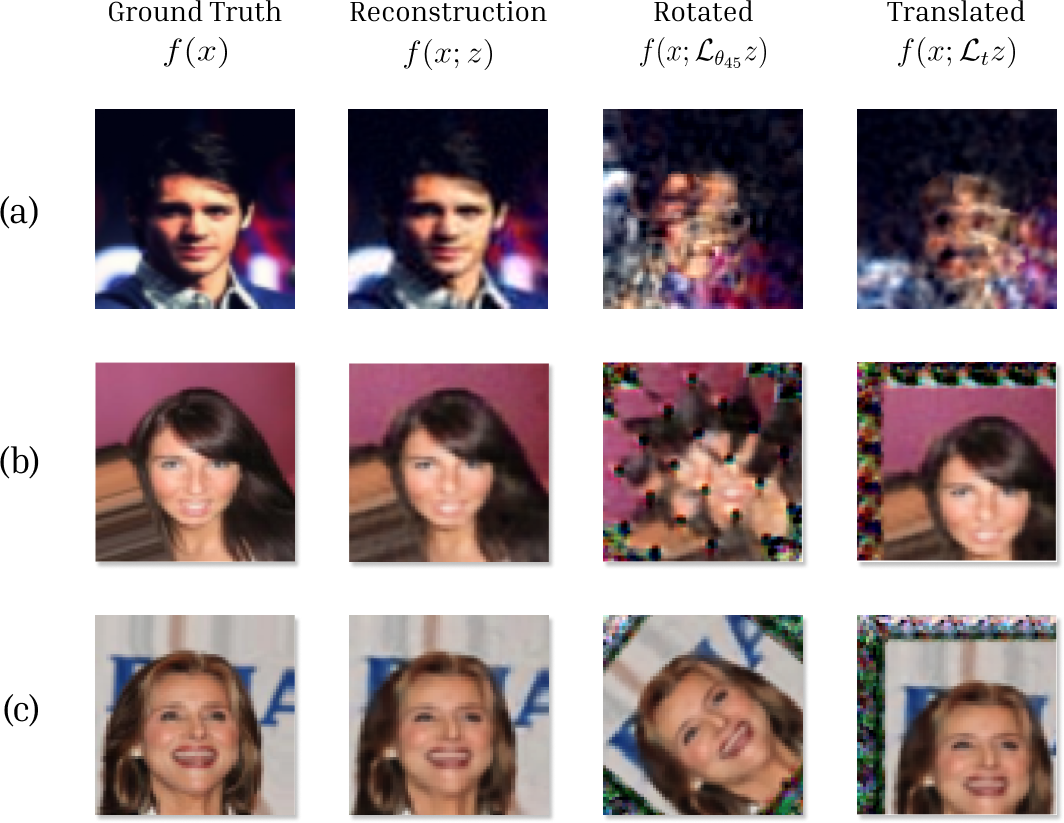}
    \caption{Transformations applied to latents $z$ for different bi-invariants $\mathbf{a}$. (a) $\mathbf{a}_{m,i}^\emptyset$ is not bi-invariant to any transformations, (b) $\mathbf{a}^{\mathbb{R}^2}_{m,i}$ is bi-invariant to translations, producing distorted patterns on rotation and (c) $\mathbf{a}_{m,i}^{\rm SE(2)}$ is bi-invariant to roto-translations; the output $f_{\theta}(x;z)$ equivaries with both rotations and translations applied to $z$.}
    \label{fig:weight-sharing}
\end{figure}

We provide visualizations for transformations applied to the latent pointclouds for different bi-invariants $\mathbf{a}$ in Fig. \ref{fig:weight-sharing}. 

\subsection{ENF with geometry-free latent sets}
To further investigate what design choices the performance of ENF results from, we provide an ablation on CIFAR10 over a geometry-free implementation of ENF. We do this by removing the pose information from the latent set, i.e. we set $z:=\{\mathbf{c}_i\}_{i=1}^N$, and use a "bi-invariant" that is only a function of $x_j$, $\mathbf{a}_{i,j}=x_j$. Since now latents do not have a position, we remove the Gaussian windowing and KNN approximation, but keep the rest of the ENF architecture as well as the hyperparameters used identical to the settings reported in Appx. \ref{appx:experimental-details} under 'CIFAR10 reconstruction and classification'. We observe highly unstable training during the reconstruction phase, and reconstruction performance on the test set converges to 22.3. We think this attributable to the fact that now any update to one of the latent codes affects the output of the NEF globally, leading to a much more complex optimization landscape. This highlights another advantage of either having a single global latent, or using locality as inductive bias; optimization of single or locally responsible latents seems to lead to a simpler optimization landscape compared to optimizing a set of global latents.

We subsequently train a simple 4 layer transformer with hidden dim 256 and 4 heads as classifier. Note that this transformer uses no positional encoding, since the latent $z$ in this setting has no associated geometry/positional information. We train for 500 epochs on the augmented dataset, after which training accuracy has converged to $95\%$. We observed overfitting early into training. Utilizing early stopping, best performance was achieved after just 5 epochs, yielding a test set accuracy of 0.47. These observations (Tab. \ref{tab:geom-free-latent}) are in line with the outcome of our other experiments; geometry-grounded latents are more informative for downstream tasks.
\begin{table}[!htb]
    \begin{minipage}{\textwidth}
    \centering
    \caption{Reconstruction PSNR (db$\uparrow$) and classification test accuracy (\%$\uparrow$) on CIFAR10 when ablating over latent geometry.}
    \label{tab:geom-free-latent}
    \begin{small}
    \scalebox{0.8}{
    \begin{tabular}{ccc}
    \toprule
     & \sc{PSNR} & \sc{ACC (\%)} \\
    \toprule
    Functa &38.1&68.3 \\
    ENF w/ pose-free latents & 22.3&47.9\\
    ENF w/ ${\mathbb{R}^2}$ latents &\textbf{42.2}&\textbf{82.1} \\
    \bottomrule
    \end{tabular}}
    \end{small}
    \end{minipage}%
\end{table} 

\subsection{Details on computational efficiency}
\label{appx:comp-effic}
To allow for more fine-grained comparison of our method with previous work, we provide details on time and memory efficiency of our approach on the CIFAR10 classification experiments listed in Tab. \ref{tab:reconstruction-classification-img} with $\mathbf{a}^{\mathbb{R}^2}$, when both Functa and ENF are fit using MAML with 3 inner loop steps. Moreover, we compare efficiency also when ablating over the KNN approximation of the attention operation. We report estimated FLOPs (obtained through JAX's AOT api), GPU memory usage per sample and training time per epoch for a batch size of 32. We see (Tab. \ref{tab:knn-experiment}) that the naive implementation that does not truncate the attention operator is significantly more FLOP-intensive and memory intensive compared to Functa \citep{dupont2022data} and the KNN approximate implementation. Functa in all settings does have considerably higher runtime, attributable to its relatively deep sequential architecture compared to the shallow single layer architecture of ENF. These results show that, besides being more performant on fine-grained downstream tasks, ENF also scales favourably compared to Functa.

\begin{table}[!htb]
    \begin{minipage}{\textwidth}
    \centering
    \caption{Computational efficiency of ENF with and without KNN approximation to the Functa baseline for the CIFAR10 experiment.}
    \label{tab:knn-experiment}
    \begin{small}
    \scalebox{0.8}{
    \begin{tabular}{cccc}
    \toprule
     & \sc{FLOPs} $(\times 10^9)$ & \sc{GPU Memory (GB/Sample)} & \sc{Time per epoch (s)} \\
    \toprule
    Functa & 28.3 & 0.61 & 2864 \\
    ENF w/o KNN approx. & 104.5 & 1.83 & 1801 \\
    ENF w KNN approx. & 22.7 & 0.40 & 207 \\
    \bottomrule
    \end{tabular}}
    \end{small}
    \end{minipage}%
\end{table} 

\subsection{Additional ShapeNet-Part segmentation results}
\label{appx:more-shapenet}
\begin{figure}[!htb]
    \centering
    \includegraphics[width=\linewidth]{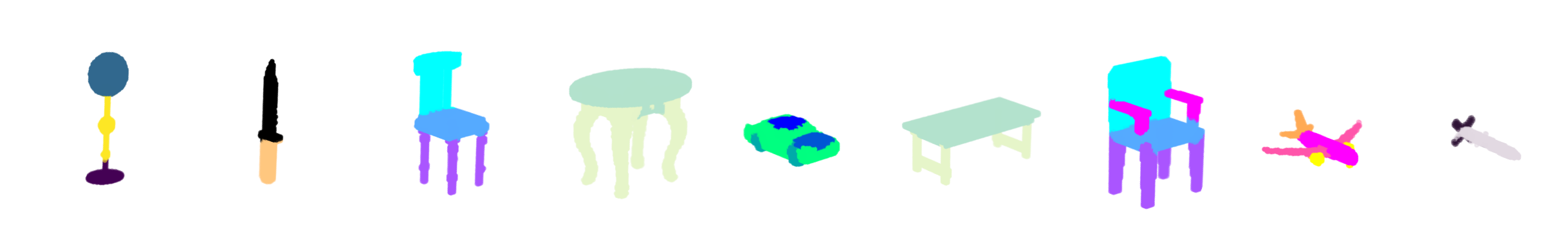}
    \caption{Qualitative examples drawn randomly from the ShapeNet segmentation test set.}
    \label{fig:shape-part-seg}
    \centering
    \captionof{table}{Segmentation class and instance averaged IOU ($\uparrow$) on ShapeNet, and mIoUs per class.}
    \label{tab:results_shape_part}
    \begin{small}
    \scalebox{0.65}{
        \begin{tabular}{lcccccccccccccccccc}
        \sc{Model} & \rotatebox[origin=c]{65}{\sc{inst mIoU}} & \rotatebox[origin=c]{65}{\sc{cls mIoU}} & \rotatebox[origin=c]{65}{\sc{airplane}}&\rotatebox[origin=c]{65}{\sc{bag}}&\rotatebox[origin=c]{65}{\sc{cap}}&\rotatebox[origin=c]{65}{\sc{car}}&\rotatebox[origin=c]{65}{\sc{chair}}&\rotatebox[origin=c]{65}{\sc{earphone}}&\rotatebox[origin=c]{65}{\sc{guitar}}&\rotatebox[origin=c]{65}{\sc{knife}}&\rotatebox[origin=c]{65}{\sc{lamp}}&\rotatebox[origin=c]{65}{\sc{laptop}}&\rotatebox[origin=c]{65}{\sc{motorbike}}&\rotatebox[origin=c]{65}{\sc{mug}}&\rotatebox[origin=c]{65}{\sc{pistol}}&\rotatebox[origin=c]{65}{\sc{rocket}}&\rotatebox[origin=c]{65}{\sc{skateboard}}&\rotatebox[origin=c]{65}{\sc{table}}\\
        \toprule
        PointNet   & 83.1 & 79.0 & 81.3 & 76.9 & 79.6 & 71.4 & 89.4 & 67.0 & 91.2 & 80.5 & 80.0 & 95.1 & 66.3 & 91.3 & 80.6 & 57.8 & 73.6 & 81.5 \\
        PointNet++ & \textbf{84.9} & \textbf{82.7} & \textbf{82.2} & \textbf{88.8} & \textbf{84.0} & \textbf{76.0} & \textbf{90.4} & 80.6 & \textbf{91.8} & 84.9 & \textbf{84.4} & 94.9 & \textbf{72.2} & \textbf{94.7} & \textbf{81.3 }& 61.1 & 74.1 & 82.3\\
        DGCNN      & 83.6 & 80.9 & 80.7 & 84.3 & 82.8 & 74.8 & 89.0 & \textbf{81.2} & 90.1 & \textbf{86.4} & 84.0 & \textbf{95.4} & 59.3 & 92.8 & 77.8 & \textbf{62.5} & 71.6 & 81.1 \\
        \midrule
        NF2vec     & 81.3 & \underline{76.9} & 80.2 & 76.2 & \underline{70.3} & 70.1 & 88.0 & \underline{65.0} & \underline{90.6} & 82.1 & 77.4 & 94.4 & 61.4 & \underline{92.7} & \underline{79.0} & \underline{56.2} & 68.6 & 78.5\\
        Functa     & \underline{82.8} & 74.8 & \underline{82.1} & 72.5 & 40.2 & 72.5 & \underline{88.5} & 60.9 & 89.2 & \underline{82.2} & 80.1 & \underline{93.9} & \underline{63.8} & 90.6 & 77.3 & 46.3 & \underline{\textbf{76.0}} & \underline{81.5}\\
        ENF        & 82.2 & 75.4 & 80.7 & \underline{77.2} & 42.1 & \underline{73.2} & 87.7 & 64.4 & 89.4 & 79.6 & \underline{80.6} & 93.8 & 62.7 & 91.8 & 76.9 & 52.5 & 74.1 & 80.6 \\
        \bottomrule
        \end{tabular}
    }
    \end{small}
    \vspace{-4mm}
\end{figure}

Below we show the full table with ShapeNet-Part segmentation results with IoUs per class in table \ref{tab:results_shape_part} and some qualitative examples in figure \ref{fig:shape-part-seg}.

\subsubsection{ShapeNet-Part segmentation without shape information}
\label{appx:shapenet-addtn}
Further investigating the results obtained in the ShapeNet Part classification task, we train an ENF $f_{\theta_\text{seg}}$ without conditioning on $z^\text{recon}$ - i.e. without any shape-specific conditioning but instead only conditioning on the object class. This model obtains class and instance mIoU of 64.3 and 69.2 respectively, indicating that a lot of points in this dataset can be correctly segmented purely based on their absolute position, and as such the backbone NeF model does not need to capture to perform decently on this dataset - though we would expect additional geometric to help with performance.

\end{document}